%% file: example_paper.tex
\documentclass{article}

\usepackage{microtype}
\usepackage{graphicx}
\usepackage{subcaption}
\usepackage{booktabs} 

\usepackage{hyperref}

\usepackage[preprint]{icml2026}

\usepackage{amsmath}
\usepackage{amssymb}
\usepackage{mathtools}
\usepackage{amsthm}
\usepackage{tabularx}
\usepackage{multirow}
\usepackage[table]{xcolor} 
\usepackage{makecell}
\usepackage{array}
\usepackage{algorithm}
\usepackage{xcolor}
\definecolor{green}{RGB}{0,255,0}
\definecolor{red}{RGB}{255,0,0}
\definecolor{blue}{RGB}{0,0,255}
\usepackage{pifont}
\usepackage{fvextra}

\fvset{breaklines=true} 
\usepackage[most]{tcolorbox}
\tcbset{
  myboxstyle/.style={
    colback=gray!20,
    colframe=gray!140,
    sharp corners,
    boxrule=0.8pt,
    fonttitle=\bfseries,
    enhanced,
    beforeafter skip=0pt,  
    left=5pt, right=5pt, top=5pt, bottom=5pt,
  }
}

\usepackage[capitalize,noabbrev]{cleveref}

\theoremstyle{plain}

\theoremstyle{definition}

\theoremstyle{remark}

\usepackage[textsize=tiny]{todonotes}

\newcommand{\method}{LaST-VLA}

\begin{document}

\twocolumn[
  \icmltitle{LaST-VLA: Thinking in Latent Spatio-Temporal Space for Vision-Language-Action in Autonomous Driving}



 \icmlsetsymbol{equal}{*}
  \icmlsetsymbol{corr}{\ding{41}}
  \icmlsetsymbol{proj}{\dag}

\begin{icmlauthorlist}
    \icmlauthor{Yuechen Luo}{equal,yyy,comp}
    \icmlauthor{Fang Li}{equal,comp}
    \icmlauthor{Shaoqing Xu}{equal,comp,proj}
    \icmlauthor{Yang Ji}{comp}
    \icmlauthor{Zehan Zhang}{comp,proj}
    \icmlauthor{Bing Wang}{comp}
    \icmlauthor{Yuannan Shen}{comp}
    \icmlauthor{Jianwei Cui}{comp}
    \icmlauthor{Long Chen}{comp}
    \icmlauthor{Guang Chen}{comp}
    \icmlauthor{Hangjun Ye}{comp,corr}
    \icmlauthor{Zhi-Xin Yang}{sch}
    \icmlauthor{Fuxi Wen}{yyy,corr}
  \end{icmlauthorlist}

  \icmlaffiliation{yyy}{Tsinghua University}
  \icmlaffiliation{sch}{University of Macau}
  \icmlaffiliation{comp}{Xiaomi EV}

  \icmlkeywords{Machine Learning, ICML}

  \vskip 0.3in
]
\printAffiliationsAndNotice{\icmlEqualContribution}




\input{sec/0_abstract}
\input{sec/1_intro}

\input{sec/2_related}
\input{sec/3_method}
\input{sec/4_exp}

\input{sec/5_conclusion}

\section*{Impact Statement}
This paper presents work whose goal is to advance the field of Machine Learning. There are many potential societal consequences of our work, none which we feel must be specifically highlighted here.

\nocite{langley00}

\bibliography{example_paper}
\bibliographystyle{icml2026}

\newpage
\appendix
\onecolumn

\input{sec/appendix}


\end{document}

%% file: sec/0_abstract.tex
\begin{abstract}
    While Vision-Language-Action (VLA) models have revolutionized autonomous driving by unifying perception and planning, their reliance on explicit textual Chain-of-Thought (CoT) leads to semantic-perceptual decoupling and perceptual-symbolic conflicts. Recent shifts toward latent reasoning attempt to bypass these bottlenecks by thinking in continuous hidden space. However, without explicit intermediate constraints, standard latent CoT often operates as a physics-agnostic representation. To address this, we propose the \textbf{La}tent \textbf{S}patio-\textbf{T}emporal \textbf{VLA}~(\textbf{\method}), a framework shifting the reasoning paradigm from discrete symbolic processing into a physically grounded Latent Spatio-Temporal CoT. By implementing a dual-feature alignment mechanism, we distill geometric constraints from 3D foundation models and dynamic foresight from world models directly into the latent space. Coupled with a progressive SFT training strategy that transitions from feature alignment to trajectory generation, and refined via Reinforcement Learning with Group Relative Policy Optimization~(GRPO) to ensure safety and rule compliance. \method~setting a new record on \textbf{NAVSIM v1} (91.3 PDMS) and \textbf{NAVSIM v2} (87.1 EPDMS), while excelling in spatial-temporal reasoning on \textbf{SURDS} and \textbf{NuDynamics} benchmarks. 
The code is available at
\href{https://github.com/luo-yc17/LaST-VLA}{LaST-VLA Code}.
\end{abstract}

%% file: sec/1_intro.tex
\section{Introduction}
\begin{figure}
    \centering
    \includegraphics[width=1.0\linewidth]{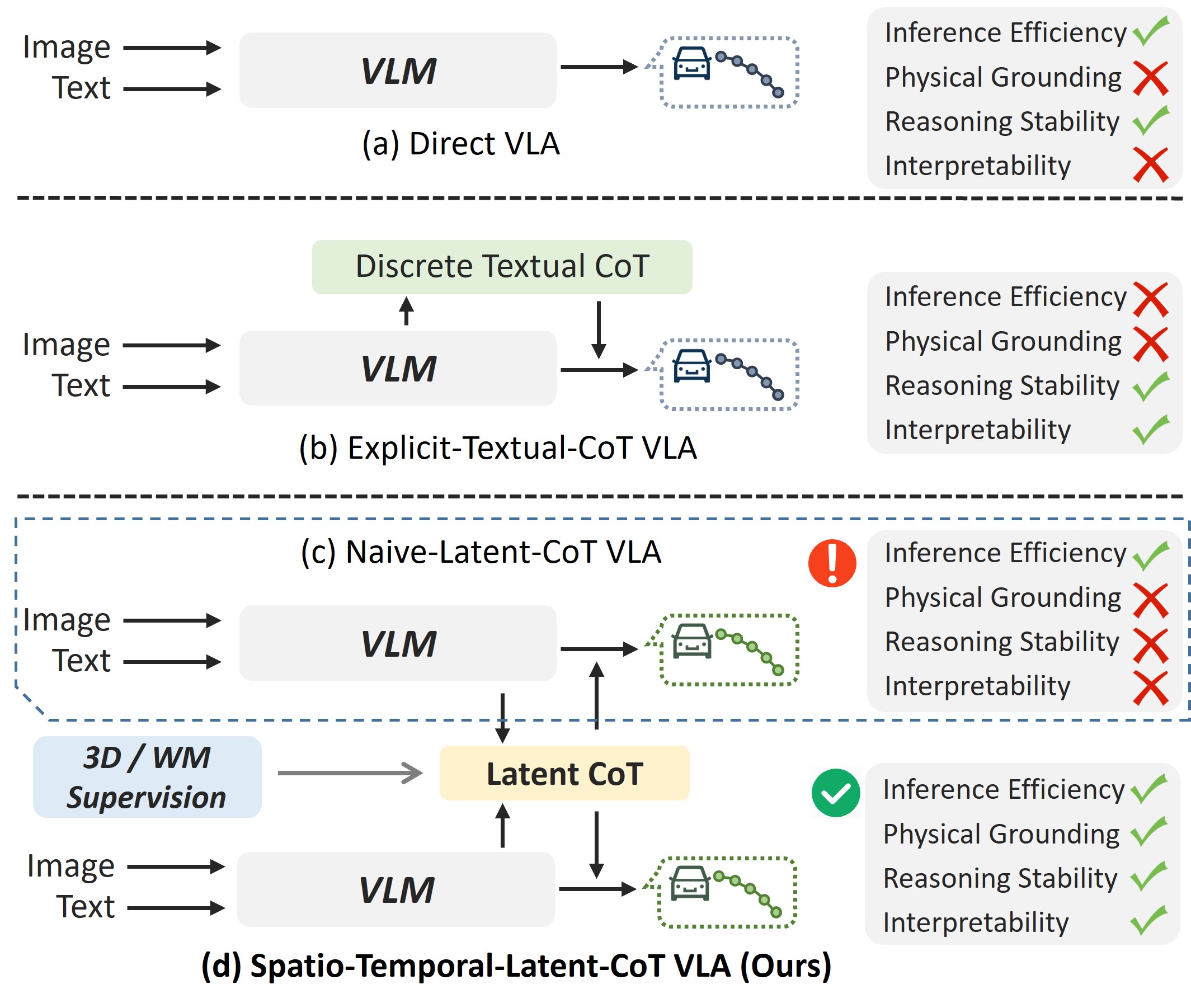}
    \caption{
    \textbf{Comparison of VLA Paradigms.} 
    \textbf{(a) Direct VLA} is efficient but lacks reasoning.
    \textbf{(b) Explicit Textual CoT} is interpretable but suffers high latency and hallucinations.
    \textbf{(c) Naive Latent CoT~(w/o supervision)} is efficient but unstable (model collapse).
    \textbf{(d) Our Spatio-Temporal Latent CoT (supervision)} aligns latent features with physical priors, achieving efficiency, stability, and grounding.
    }
    \label{fig:intro}
    \vskip -2.5em
\end{figure}

The evolution of autonomous driving is undergoing a profound shift: from traditional modular pipelines~\cite{hu2023planning, zhang2024sparsead, jiang2023vad} to Vision-Language-Action (VLA) models~\cite{wang2025omnidrive, li2025recogdrive, wang2025alpamayo}. While modular approaches offer interpretability, they often struggle with complex and long-tail scenarios due to rigid rule-based designs and error propagation across modules. In contrast, VLA models unify visual perception and driving policy into a holistic system, demonstrating superior scene generalization and natural language instruction following capabilities.

While VLA models have demonstrated promising performance surpassing traditional end-to-end (E2E) approaches. They remain hindered by fundamental limitations that impede their deployment in real world. Early works~\cite{wang2025omnidrive, fu2025orion} directly generate trajectories without explicit intermediate reasoning, as shown in \cref{fig:intro}(a), suffering from poor interpretability. Recent efforts~\cite{hwang2024emma, li2025drive} unlock strong reasoning capabilities through explicit Chain-of-Thought (CoT), thereby achieving interpretability and enabling the VLA to solve complex scenarios in step-by-step manner, as shown in \cref{fig:intro}(b). {However, projecting dense visual data into discrete text creates an inherent semantic gap, often leading to hallucinations inconsistent with visual input. Consequently, the planner often disregards visual evidence and instead follows flawed linguistic guidance, leading to potentially hazardous decision failures.} Moreover, the generation of extensive intermediate sequences increases inference cost and results in redundant over-thinking, as discovered in AdaThinkDrive~\cite{adathinkdrive} and AutoVLA
~\cite{zhou2025autovla}.
In response, recent methods~\cite{hao2024training, wei2025sim} argue that natural language may not always be the optimal medium for structured reasoning. They shift from explicit textual reasoning to implicit latent space reasoning, as shown in \cref{fig:intro}(c), achieving higher computational efficiency by bypassing the generation of lengthy textual chains. However, our empirical findings indicate that these methods typically rely solely on supervision from the final answer, without intermediate constraints. This absence of intermediate guidance leads to an unstable training process and renders the model susceptible to collapse.

To address these challenges, we propose LaST-VLA, a novel latent reasoning framework. Inspired by World Action Models~\cite{xia2025drivelaw} and VLM methods with spatial forcing~\cite{li2025spatial, zheng2025learning}, we leverage latent representations from geometry foundation models and video world models as supervision targets, as shown in \cref{fig:intro}(d). The design paradigm enables the model to receive temporally stable, step-level supervisory signals while simultaneously embedding spatial-temporal perceptual capabilities into latent reasoning process.

Our methodology employs a progressive training strategy to ensure the model effectively internalizes these grounded reasoning capabilities. 
As illustrated in the top-right corner of \cref{fig:overview}, the first stage of supervised learning aims to equip the model with spatial-temporal reasoning capabilities, while the second stage focuses on enabling the model to learn specific planning tasks.
Furthermore, we employ Reinforcement Learning (RL) to enhance specific driving decision-making capabilities, as shown in the bottom-right corner of \cref{fig:overview}. Following SFT, the model is equipped with spatial-temporal understanding and planning abilities. Subsequently, we refine the model using RL based on metrics such as driving safety and comfort, ultimately endowing it with superior driving proficiency.

In summary, our key contributions are:
\begin{itemize}
\vspace{-1em}
\item We highlight two critical deficiencies in existing VLA-based autonomous driving: the disconnect between language and physical reality, and the safety hazards induced by excessive reliance on linguistic priors.
\vspace{-0.5em}

\item We propose \method, which unifies instruction following and dynamic prediction via a latent spatio-temporal CoT. This design surpasses both direct explicit textual reasoning and unsupervised latent approaches, overcoming the precision limitations of the former and the training instability of the latter.
\vspace{-0.5em}

\item We design a progressive training strategy that first equips the model with spatial-temporal understanding capabilities, and subsequently enables it to acquire planning. Experiments demonstrate that our method significantly outperforms state-of-the-art baselines across multiple autonomous driving benchmarks.
\end{itemize}

%% file: sec/2_related.tex
\begin{figure*}[htb]
    \centering
    \includegraphics[width=0.95\linewidth]{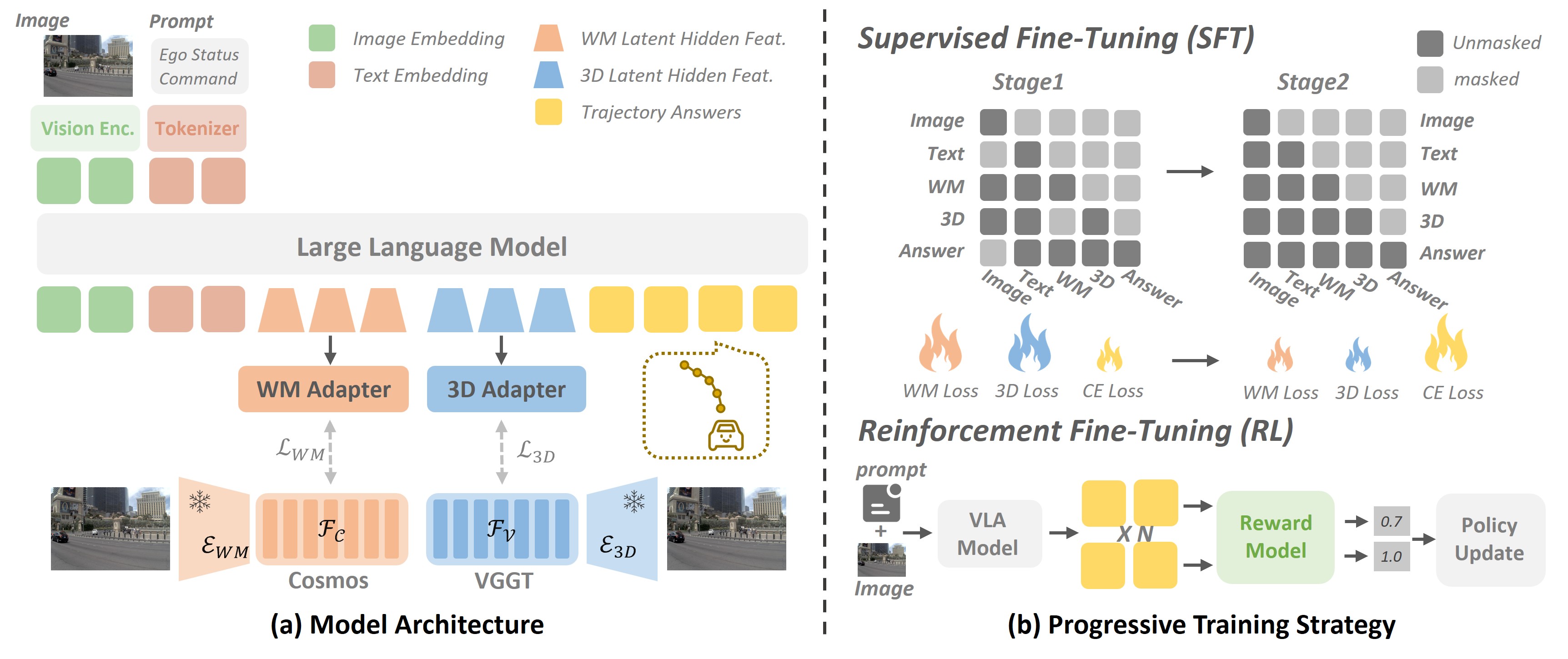}
    \caption{\textbf{Overview of the \method~framework.} 
    \textbf{(a) Model Architecture:} The model constructs a Latent CoT by aligning hidden states with dynamic and geometric priors distilled from foundation models (Cosmos and VGGT) via specialized adapters.
    \textbf{(b) Progressive Training Strategy:} The pipeline features a two-stage SFT phase that utilizes structured causal masking to enforce physically grounded reasoning, followed by RL fine-tuning to directly optimize the policy for driving safety and compliance.}
    \label{fig:overview}
    \vskip -1em
\end{figure*}

\section{Related Work}

\subsection{VLA models in Autonomous Driving}
Traditional E2E methods~\cite{jiang2023vad, hu2023planning} rely on modular pipelines but often lack natural interfaces for high-level intent. In contrast, VLAs unify these stages, demonstrating superior contextual understanding. Early research focused on scene understanding and meta-action generation~\cite{tian2024drivevlm, jiang2025alphadrive}. Some approaches employ VLMs directly for textual trajectory prediction. For instance, Orion~\cite{fu2025orion} and OmniDrive~\cite{wang2025omnidrive} adopt StreamPETR~\cite{wang2023exploring} as Q-Former3D to compress scene features. This architecture bridges the visual reasoning space and facilitates subsequent textual trajectory prediction. Similarly, EMMA~\cite{hwang2024emma}, trained on large-scale datasets, leverages Gemini~\cite{team2023gemini} to predict discrete textual perception and planning outputs. Others integrate VLMs with E2E models in fast-slow systems: DriveVLM~\cite{sima2024drivelm} uses VLMs for coarse trajectory prediction, which is refined by an E2E model. More recent methods introduce a textual cot prior to trajectory prediction, leveraging the common sense reasoning capabilities of LLM backbones to enhance planning performance, particularly in rare or complex scenarios~\cite{hwang2024emma, wang2025alpamayo, wang2024drivecot}.

\subsection{Latent Chain-of-Thought}
Textual CoT has become a popular strategy for eliciting reasoning in VLMs and VLAs. However, some works~\cite{zhu2026analyzing} revealed that visual information is suppressed by the generated text, leading the answer to no longer focus on it. Moreover, recent research~\cite{lou2025adacot, zhang2025adaptthink} discovered that a large amount of meaningless CoT also leads to low efficiency and accuracy. In autonomous driving, AdaThinkDrive~\cite{adathinkdrive} and AutoVLA~\cite{zhou2025autovla} showed that explicit textual CoT can hinder planning performance in some scenarios. To overcome the shortcomings of the textual CoT, some works~\cite{cheng2024compressed, hao2024training} begun to explore latent reasoning in LLMs, where intermediate computations are performed in continuous latent space rather than in textual space. This paradigm enables more cost-effective inference budget. 
Building on these ideas, subsequent works~\cite{ray2025mull, li2025latent} extend latent reasoning to VLMs, achieving latent spatial reasoning. 
In contrast, we observe that leaving the continuous latent CoT unsupervised degrades planning performance. To address this, we propose \textbf{Think with Latent Spatio-Temporal}, a mechanism that leverages geometric and dynamic priors to supervise the latent CoT, thereby enhancing the fidelity and robustness of trajectory planning.

%% file: sec/3_method.tex
\section{Method}

In this section, we present the  \method~framework, as illustrated in Figure~\ref{fig:overview}. This architecture is designed to bridge the gap between perception and planning via a Latent Spatio-Temporal CoT mechanism. The overall pipeline comprises three core components: (1) Latent Spatio-Temporal CoT, (2) Progressive Two-Stage SFT Strategy, and (3) Latent-Grounded Trajectory Refinement via GRPO.

\subsection{Preliminaries}
\label{sec:problem_formulation}

In this work, we formulate end-to-end autonomous driving planning as a conditional generation problem augmented by latent reasoning.

\noindent \textbf{Inputs.}~~
At any given time step $t$, the system receives a multimodal query tuple $\mathbf{Q}_t = (I_t, L)$, comprising the front-view camera image $I_t \in \mathbb{R}^{H \times W \times 3}$ and the text input $L$. Specifically, $L$ includes the navigation instruction $l$ (e.g., \textit{``Turn left''}), the ego-vehicle state vector $s_{\text{ego}} \in \mathbb{R}^{d_s}$ (velocity, acceleration), and historical trajectory $h_{\text{his}} \in \mathbb{R}^{T \times 3}$.

\noindent \textbf{Latent Reasoning.}~~
Deviating from standard VLA paradigms that directly map inputs to actions, we introduce a \textit{Latent Spatio-Temporal CoT}, denoted as $H$, to serve as a continuous reasoning bridge. Derived from the final hidden states of the VLM backbone, we explicitly structure $H$ into two decoupled features: $H = \{H^{\text{dyn}}, H^{\text{geo}}\}$. Here, $H^{\text{dyn}}$ captures temporal  dynamics evolution~(WM), while $H^{\text{geo}}$ encodes spatial geometry~(3D). These continuous latent variables are autoregressively generated to explicitly ground the reasoning process in physical properties.

\noindent \textbf{Probabilistic Modeling.}~~
Our objective is to learn a hierarchical model that generates the trajectory $\mathbf{a}_t$ conditioned on the latent CoT $H$. We formalize this by decoupling the joint distribution into a thinking phase and a planning phase:
\begin{equation}
    P_{\theta}(\mathbf{a}_t, \mid \mathbf{Q}_t) = \underbrace{P(H^{\text{dyn}}, H^{\text{geo}} \mid \mathbf{Q}_t)}_{\text{Thinking Process}} \cdot \underbrace{P(\mathbf{a}_t \mid H^{\text{dyn}}, H^{\text{geo}}, \mathbf{Q}_t)}_{\text{Planning Process}}.
\end{equation}

In this formulation, the model first acts as a \textit{Thinker} to generate the continuous reasoning states $H$, and subsequently functions as a \textit{Planner} to predict trajectory waypoints $\mathbf{a}_t$ conditioned on these grounded thoughts. Notably, both probability distributions are parameterized by the same VLA, realized through unified auto-regressive process.

\subsection{Latent Spatio-Temporal CoT}
\label{sec:latent_cot}

 To teach the model to reason with 3D geometry and world dynamics, conventional approaches typically rely on explicit reconstruction targets, such as predicting dense depth maps or future video frames. However, these methods suffer from high computational overhead and information redundancy, as they compel the model to focus on irrelevant textural details rather than critical physical states. Instead, as depicted in Figure~\ref{fig:overview}~(a), we introduce external foundation models as feature-level teachers during training, distilling their structured knowledge into the continuous latent tokens generated from the VLM's reasoning process. This strategy facilitates effective physical understanding without the burden of pixel-level generation.

Specifically, the reasoning process begins with visual encoding. Given the input image $I_t$, vision encoder $\mathcal{V}$ is employed to extract patch-level visual embeddings $E_{\text{img}} = \mathcal{V}(I_t) \in \mathbb{R}^{N \times D}$. These visual features are concatenated with the embeddings of linguistic instructions $E_{L}$ to form the multimodal input sequence. The VLM  $\pi_{\theta}$ autoregressively generates the hidden states. Let $h_k$ denote the output of the last layer at step $k$. The latent CoT sequence $H$ is derived as:
\begin{equation}
    H = \{h_k\}_{k=1}^{K} = \pi_{\theta}(E_{\text{img}}, E_{\text{L}}),
\end{equation}
where $K$ is the length of the reasoning chain. This sequence is strictly partitioned into dynamic ($H^{\text{dyn}}$) and geometric ($H^{\text{geo}}$) streams to facilitate two specialized physical alignments with adapters. To prevent the adapter from shortcutting alignment by attending solely to raw pixel patterns, we apply a random binary mask $M$ to the visual embeddings, yielding $\tilde{E}_{\text{img}} = E_{\text{img}} \odot M$. These masked features and continuous CoT hidden states are as input to adapters for alignment with different priors.

\begin{figure}
    \centering
    \includegraphics[width=1.0\linewidth]{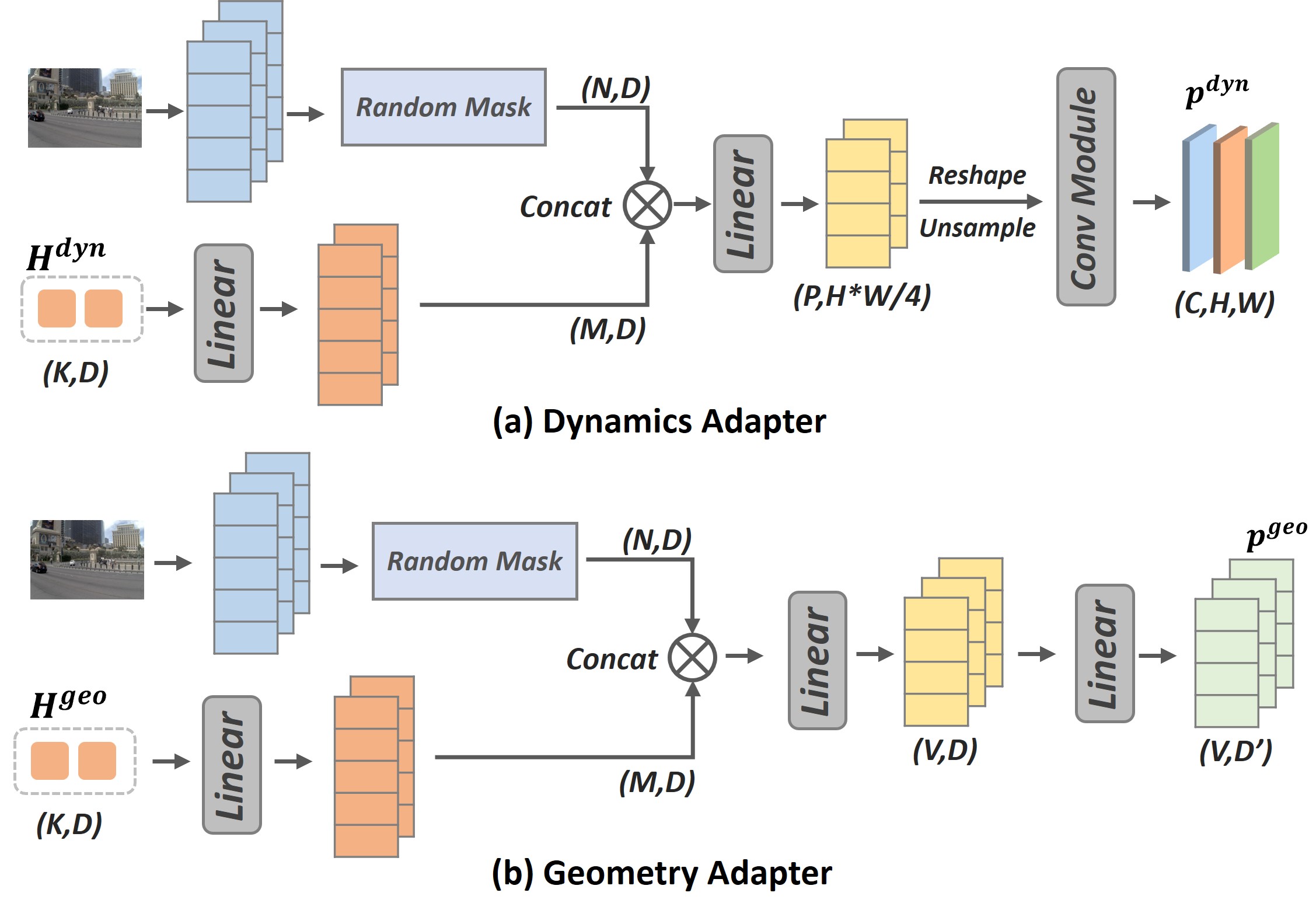}
    \vskip -0.5em
    \caption{\textbf{Architecture of the Dynamics (a) and Geometry (b) Adapters.} Random mask is used only during training.}
    \label{fig:adapter}
    \vskip -1em
\end{figure}

\noindent\textbf{Dynamics Adapter} ($\Phi_{\text{dyn}}$). As illustrated in Figure~\ref{fig:adapter}(a), this adapter bridges the sequential hidden states with the representation space of video world model (Cosmos~\cite{agarwal2025cosmos}). Unlike the static nature of standard linguistic tokens, the world model's latent space inherently encodes temporal dynamics evolution. 
Consequently, by projecting the linear token sequence to align with this dynamic manifold, $\Phi_{\text{dyn}}$ effectively captures future motion prior of traffic participants and continuous environmental changes.

\noindent\textbf{Geometry Adapter} ($\Phi_{\text{geo}}$). As illustrated in Figure~\ref{fig:adapter}(b), this adapter aligns the spatial hidden states with dense feature space of 3D foundation model of VGGT~\cite{wang2025vggt}. By fusing the linguistic latent states $H^{\text{geo}}$ with the original visual embeddings $\tilde{E}_{\text{img}}$, it recovers metrically accurate spatial priors, such as scene depth and occupancy structures, directly in the latent space.
The overall transformation process is formally defined as:
\begin{equation}
    {\mathbf{p}}^{\text{geo}} = \Phi_{\text{geo}}(H^{\text{geo}}, \tilde{E}_{\text{img}}), \quad
    {\mathbf{p}}^{\text{dyn}} = \Phi_{\text{dyn}}(H^{\text{dyn}}, \tilde{E}_{\text{img}}),
    \label{equation:adapter}
\end{equation}
The latent CoT serves as an implicit reasoning bridge that encodes what the world looks like in 3D space and how it will change, while remaining fully differentiable and aligned with the downstream trajectory output space.

\noindent\textbf{Latent-Conditioned Planning.}
Finally, the grounded latent chain-of-thought is integrated in the trajectory generation process. Following the latent reasoning phase, our policy autoregressively predicts the future waypoints $\mathbf{a}_t$ (represented as textual tokens) conditioned on these physical reasonings:
\begin{equation}
    \mathbf{a}_t \sim \pi_{\theta}(\mathbf{a}_t \mid {H}^{\text{dyn}}, {H}^{\text{geo}}, \mathbf{Q}_t).
\end{equation}
This ensures that the final planning decision is explicitly driven by the spatio-temporal understanding distilled in the latent space.

\subsection{Progressive Two-Stage SFT Strategy}
\label{sec:training_strategy}

As illustrated in Figure~\ref{fig:overview}~(b), our method introduces external foundation models as feature teachers during training, and distills their structured knowledge into the continuous latent tokens generated in the VLM's reasoning process. To implement this effectively, we propose a \textit{Progressive Two-Stage Supervised Fine-Tuning strategy} that decouples the complex objective into learning to \textit{think} and learning to \textit{action}.

\noindent\textbf{Unified Optimization Objective.}~~
We first formulate the unified training objective for both stages. To ground the latents in physical reality, we employ feature distillation from frozen foundation models, denoting $F_{\text{Cosmos}}$ as the dynamic feature derived from the Cosmos and $F_{\text{VGGT}}$ as the geometric features from the VGGT aggregator. For dynamic representation, we designate three sets of latent tokens to align with the temporal features of $F_{\text{Cosmos}}$. These tokens encode dynamics across progressive temporal scales, capturing short-term, medium-term, and long-term motion. Simultaneously, we allocate a distinct set of latent tokens to align with $F_{\text{VGGT}}$ to capture static geometric constraints. The alignment losses are calculated as the Mean Squared Error (MSE) between the adapter-projected features $\mathbf{p}$ and these teacher targets:
\begin{equation}
    \mathcal{L}_{\text{WM}} = \|\mathbf{p}^{\text{dyn}} - F_{\text{Cosmos}}\|_2^2, 
    \mathcal{L}_{\text{3D}} = \|\mathbf{p}^{\text{geo}} - F_{\text{VGGT}}\|_2^2,
\end{equation}
The total SFT loss combines the trajectory prediction error with these physical alignment terms:
\begin{equation}
    \mathcal{L}_{\text{total}} = \lambda_{\text{action}} \mathcal{L}_{\text{CE}} + \lambda_{\text{WM}} \mathcal{L}_{\text{WM}} + \lambda_{\text{3D}} \mathcal{L}_{\text{3D}}.
\end{equation}
where $\mathcal{L}_{\text{CE}}$ denotes the cross-entropy loss for action generation, and $\lambda$ are the balancing coefficients adjusted dynamically across stages.

\noindent\textbf{Phase I: Physics-Aware Alignment.}~~
In this initial phase, we prioritize learning physical knowledge over trajectory generation. We set the loss weights to $\lambda_{\text{WM}} = \lambda_{\text{3D}} = 1.0 \gg \lambda_{\text{action}} = 0.01$. This forces the latent CoT $H$ to strictly align with the teacher models' geometric and dynamic representations. 
To ensure the planner relies on this reasoning, we apply a \textbf{Structured Causal Masking}~(Figure~\ref{fig:overview}(b)): (1) \textbf{Latent Mutual Masking:} We mask the 3D and WM tokens from each other so they learn independently. (2) \textbf{Visual Bottleneck Masking:} We prevent the action tokens from attending to the raw image embeddings $E_{\text{img}}$. This forces the model to compress all necessary visual information into $H$, making the latent thoughts the only information bridge for decision-making.

\noindent\textbf{Phase II: Latent-Grounded Planning.}~~
Once the reasoning capability is established, we switch to the second phase to refine the driving policy. We invert the weights to $\lambda_{\text{action}} = 1.0 \gg \lambda_{\text{WM}} = \lambda_{\text{3D}} = 0.01$, prioritizing accurate trajectory prediction. 
In this stage, we allow action tokens to attend to both the latent CoT $H$ and the raw image embeddings $E_{\text{img}}$. This allows the model to combine two types of information: the high-level physical understanding from $H$ and the fine-grained visual details from the original image. The reduced alignment weights maintain consistency in the reasoning, while the planner learns to use both signals for robust driving.

\subsection{Latent-Grounded Trajectory Refinement via GRPO}
\label{sec:rl_refinement}

Following the progressive SFT phase, our policy has acquired robust spatio-temporal capabilities, where the geometric and dynamic latent CoT ground the reasoning process. To further elevate the policy's execution capability, we freeze the dynamics and geometry adapters. We employ Group Relative Policy Optimization (GRPO)~\cite{shao2024deepseekmath} to optimize the VLA's action generation by maximizing trajectory-level rewards, using established latent reasoning as stable internal guidance.

\noindent \textbf{Reward Formulation.}~~
To incentivize the model to generate safe, compliant, and precise driving behaviors, we design a composite reward function $R$ comprising three distinct components. \textbf{PDMS Reward} ($R_{\text{traj}}$) evaluates the overall quality of the predicted trajectory using the Predictive Driver Model Score~\cite{dauner2024navsim}, normalized to a continuous value in $[0, 1]$. \textbf{Format Reward} ($R_{\text{fmt}}$) is a discrete indicator that strictly penalizes adherence failures to the required output structure. \textbf{Goal Reward} ($R_{\text{goal}}$) encourages endpoint precision by assigning tiered incentives based on the $L_1$ distance between the predicted and ground-truth endpoints. Calculation details are provided in the Appendix~\ref{appendix:reward}. The total reward for a trajectory is integrated as:
\begin{equation}
    R = \lambda_{traj} R_{\text{traj}} + \lambda_{fmt} R_{\text{fmt}} + \lambda_{goal}R_{\text{goal}}.
\end{equation}

\noindent \textbf{Optimization Objective.}~~
We employ GRPO as the reinforcement learning algorithm to optimize the policy $\pi_{\theta}$. For each input query $q$, we sample a group of $G$ candidate outputs $\{o_1, o_2, \dots, o_G\}$ from the sampling policy $\pi_{\theta_{old}}$. The optimization process leverages the relative advantage of these outputs to update the policy, incorporating a clipped objective to ensure training stability and a KL-divergence term to prevent excessive deviation from the reference policy $\pi_{\text{ref}}$. The objective function is formulated as follows:
\begin{equation}
    \mathcal{J}(\theta) = \mathbb{E}_{q \sim D, \{o_i\} \sim \pi_{\theta_{old}}} \left[ \frac{1}{G} \sum_{i=1}^G \mathcal{J}_i - \beta \mathbb{D}_{KL}(\pi_{\theta} || \pi_{\text{ref}}) \right],
\end{equation}
\vskip -1em
\begin{equation}
    \mathcal{J}_i = \min \left( c_i A_i, \text{clip}\left(c_i, 1 - \epsilon, 1 + \epsilon\right) A_i \right).
\end{equation}

Here, $c_i=\pi_\theta(o_i | q) / \pi_{\theta_{old}}(o_i | q)$, $\epsilon$ and $\beta$ are hyperparameters controlling the clipping range and regularization strength, respectively. Advantage $A_i$ is computed by standardizing the rewards within the group: $A_i = (R_i - \text{mean}(R)) / \text{std}(R)$, where $R_i$ is the reward for output.

%% file: sec/4_exp.tex
\section{Experiment} \label{exp}

\begin{table*}[t]
\renewcommand{\arraystretch}{1}
\begin{center}
\caption{Comparison with state-of-the-art methods on NAVSIMv1 with PDMS. * indicates models utilizing Textual CoT.}
\begin{tabularx}{\textwidth}{l|c|c|XXXXX|c}
\toprule
\textbf{Method} & \textbf{Image} & \textbf{Lidar} & \textbf{NC}$\uparrow$ & \textbf{DAC}$\uparrow$ & \textbf{TTC}$\uparrow$ & \textbf{CF}$\uparrow$ & \textbf{EP}$\uparrow$ & \textbf{PDMS}$\uparrow$ \\
\midrule
Constant Velocity & & & 68.0 & 57.8 & 50.0 & \textbf{100} & 19.4 & 20.6 \\
Ego Status MLP & & & 93.0 & 77.3 & 83.6 & \textbf{100} & 62.8 & 65.6 \\
\midrule
UniAD~\cite{hu2023planning} & \ding{51} & & 97.8 & 91.9 & 92.9 & \textbf{100} & 78.8 & 83.4 \\
DiffusionDrive~\cite{liao2025diffusiondrive} & \ding{51} & \ding{51} & 98.2 & 96.2 & 94.7 & \textbf{100} & 82.2 & 88.1 \\
WoTE~\cite{li2025end} & \ding{51} & \ding{51} & 98.5 & 96.8 & 94.9 & 99.9 & 81.9 & 88.3 \\
DriveVLA-W0-7B~\cite{li2025drivevla} & \ding{51} &  & \textbf{98.7} & \textbf{99.1} & 94.9 & 99.3 & 83.3 & 90.2 \\
AdaThinkDrive-8B*~\cite{adathinkdrive} & \ding{51} &  & 98.4 & 97.8 & 95.2 & \textbf{100} & 84.4 & 90.3 \\
Recogdrive-2B~\cite{li2025recogdrive} & \ding{51} &  & 97.9 & 97.3 & 94.9 & \textbf{100} & \textbf{87.3} & 90.8 \\
\midrule  
 InternVL3-8B~(SFT)* & \ding{51} & & 98.3 & 92.3 & 94.7 & \textbf{100} & 77.2 & 83.9 \\
 InternVL3-8B~(RL)* & \ding{51} & & 98.3 & 94.0 & 94.7 & \textbf{100} & 83.0 & 87.2  \\
\midrule
  \rowcolor{gray!30} \textbf{\method-2B}~(SFT) & \ding{51} & & 98.5 & 95.2 & \textbf{95.6} & \textbf{100} & 80.4 & 87.0 \\
 \rowcolor{gray!30} \textbf{\method-2B}~(RL) & \ding{51} & & 98.6 & 97.7 & \textbf{95.6} & \textbf{100} & 86.5 & 91.1 \\
 \rowcolor{gray!30} \textbf{\method-8B}~(SFT) & \ding{51} & & \textbf{98.7} & 95.4 & \textbf{95.6} & \textbf{100} & 80.5 & 87.3  \\
 \rowcolor{gray!30} \textbf{\method-8B}~(RL) & \ding{51} & & \textbf{98.7} & 97.9 & \textbf{95.6} & \textbf{100} & 86.8 & \textbf{91.3} \\
\bottomrule
\end{tabularx}
\label{table:main_table}
\vskip -1em
\end{center}
\end{table*}

\begin{table*}[t!]
\centering
\vskip -0.5em
\caption{Comparison with state-of-the-art methods on NAVSIMv2 with EPDMS. 
}
\label{table:navsimv2}
\resizebox{\textwidth}{!}{
\begin{tabular}{l|ccccc|cccc|c} 
\toprule
\textbf{Method} & \textbf{NC $\uparrow$} & \textbf{DAC $\uparrow$} & \textbf{DDC $\uparrow$} & \textbf{TLC $\uparrow$} & \textbf{EP $\uparrow$} & \textbf{TTC $\uparrow$} & \textbf{LK $\uparrow$} & \textbf{HC $\uparrow$} & \textbf{EC $\uparrow$} & \textbf{EPDMS $\uparrow$} \\
\midrule
Ego Status & 93.1 & 77.9 & 92.7 & 99.6 & 86.0 & 91.5 & 89.4 & \textbf{98.3} & 85.4 & 64.0 \\
TransFuser~\cite{chitta2022transfuser} & 96.9 & 89.9 & 97.8 & 99.7 & 87.1 & 95.4 & 92.7 & \textbf{98.3} & 87.2 & 76.7 \\
DiffusionDrive~\cite{liao2025diffusiondrive} & 98.2 & 95.9 & 99.4 & \textbf{99.8} & 87.5 & 97.3 & \textbf{96.8} & \textbf{98.3} & \textbf{87.7} & 84.5 \\
HydraMDP++~\cite{li2024hydra} & 98.5 & 98.5 & \textbf{99.5} & 99.7 & 87.4 & 97.9 & 95.8 & 98.2 & 75.7 & 85.6 \\
DriveVLA-W0-7B~\cite{li2025drivevla} & 98.5 & \textbf{99.1} & 98.0 & 99.7 & 86.4 & 98.1 & 93.2 & 97.9 & 58.9 & 86.1 \\
\midrule
\rowcolor{gray!30} \textbf{\method-2B}~(RL) & \textbf{98.6} & 97.7 & 99.1 & 99.7 & 90.2 & \textbf{98.2} & 96.6 & \textbf{98.3} & 86.8 & 86.8 \\
\rowcolor{gray!30} \textbf{\method-8B}~(RL) & \textbf{98.7} & 97.9 & 99.2 & 99.7 & \textbf{90.3} & \textbf{98.2} & 96.6 & \textbf{98.3} & 86.3 & \textbf{87.1} \\
\bottomrule
\end{tabular}
}
\vskip -1.0em
\end{table*}

\subsection{Implementation details}
\noindent \textbf{Datasets.} 
We primarily utilize \textbf{NAVSIM}~\cite{dauner2024navsim}, a planning-oriented benchmark derived from OpenScene. From the standard 85k split, we curate a 24k subset of challenging scenarios, denoted as navtrain-hard-24k, to enhance training efficiency. Additionally, we employ \textbf{SURDS}~\cite{drivemllm}, built on nuScenes, to evaluate 3D spatial reasoning capabilities. To further assess dynamic scene understanding, we constructed the \textbf{NuDynamics} benchmark following SURDS~(details in Appendix~\ref{appendix:nudynamics}).

\noindent \textbf{Metrics.}\label{metric} 
\label{sec:metric}
We evaluate method across two primary dimensions: trajectory planning and spatial-dynamic reasoning.

For planning evaluation, we leverage the Predictive Driver Model Score (PDMS) for \textbf{NAVSIMv1}~\cite{dauner2024navsim} and the Extended Predictive Driver Model Score (EPDMS) for \textbf{NAVSIMv2}~\cite{cao2025pseudo} as the closed-loop planning metrics~(details in Appendix~\ref{appendix:metric}). 

For spatial and dynamic scene reasoning, we employ two benchmarks. On SURDS, we report accuracy across its key tasks, including Yaw Angle Determination~(Yaw), Pixel Location Estimation~(Pixel), Depth Range Determination~(Depth), Distance Estimation~(Dis), Left/Right Determination~(L/R), and Front/Behind Determination(F/B). On NuDynamics, we utilize the Motion State Estimation~(Motion) metric to assess dynamic understanding.

\noindent \textbf{Training Details.}
We utilize InternVL3~\cite{zhu2025internvl3} as the foundation model, trained across: SFT and RL phases. In the first SFT stage, we fine-tune the model on navtrain-hard-24k for 2 epochs. In the second SFT stage, the model is trained on the full navtrain dataset for 2 epochs. Finally, during the RL phase, we train the model on navtrain-hard-24k for 2 epochs. For SURDS and NuDynamics, we only perform SFT for 2 epochs. Additional details and hyperparameters are provided in the Appendix~\ref{appendix:param}.

\begin{table*}[t!]
\centering
\small
\renewcommand{\arraystretch}{1.0}
\setlength{\tabcolsep}{8.8pt}
\caption{\textbf{Comparison with state-of-the-art methods on the SURDS and NuDynamics benchmarks.} \textbf{Score} represents the average performance on SURDS. \textbf{Motion} denotes the Motion State Estimation accuracy on NuDynamics. * indicates models trained with SFT+RL, and $\dagger$ indicates models trained with Only SFT. All methods except \method~utilize Textual CoT.}
\begin{tabular}{l|ccc|ccc|c|c}
\toprule
\multirow{2}{*}{\textbf{Method}} & \multicolumn{3}{c|}{\textbf{Single-object}(\%)$\uparrow$} & \multicolumn{3}{c|}{\textbf{Multi-object(\%)$\uparrow$}} & \multirow{2}{*}{\textbf{Score(\%)$\uparrow$}} & \multirow{2}{*}{\textbf{Motion}(\%)$\uparrow$} \\
\cmidrule{2-7}
& \textbf{Yaw} & \textbf{Pixel} & \textbf{Depth} & \textbf{Dis} & \textbf{L/R} & \textbf{F/B} & & \\
\midrule
Qwen2.5-VL-72B-Instruct & 11.57 & 6.13 & 44.00 & 58.05 & 66.16 & 14.92 & 33.47 & 54.79 \\
Qwen2.5-VL-7B-Instruct & 7.57 & 3.46 & 25.95 & 11.46 & 17.95 & 9.30 & 12.61 & 48.77 \\
Qwen2.5-VL-3B-Instruct & 6.27 & 3.81 & 27.68 & 17.84 & 14.81 & 10.49 & 13.48 & 35.34 \\
\midrule
SURDS-3B*~\cite{drivemllm} & 20.97 & 44.81 & 69.84 & 49.30 & 51.35 & 8.54 & 40.80 & - \\
InternVL3-2B$\dagger$ & 37.30 & 48.90 & 99.90 & 80.54 & 82.81 & 77.95 & 71.23 & 66.54 \\
InternVL3-8B$\dagger$ & 45.19 & 61.83 & \textbf{100.00} & 82.05 & 86.16 & 84.43 & 76.61 & 73.97 \\
\midrule
\rowcolor{gray!30} \textbf{\method-2B}$\dagger$ & 60.50 & 54.10 & 100.00 & 84.32 & 89.24 & 88.22 & 79.40 & 71.80 \\
\rowcolor{gray!30} \textbf{\method-8B}$\dagger$ & \textbf{70.16} & \textbf{71.28} & \textbf{100.00} & \textbf{86.05} & \textbf{90.27} & \textbf{88.00} & \textbf{84.29} & \textbf{81.19} \\
\bottomrule
\end{tabular}
\label{table:drivemllm}
\end{table*}

\subsection{Performance Comparison}

\noindent \textbf{NAVSIM Benchmark.} 
Table~\ref{table:main_table} presents the performance comparison on NAVSIMv1 benchmark. \method-8B~achieves a PDMS of 91.3, setting a new state-of-the-art (SOTA). This result surpasses the previous best vision-only method~(Recogdrive-2B) by 0.5 PDMS. Remarkably, even our lightweight Last-VLA-2B delivers exceptional performance with 91.1 PDMS, outperforming Recogdrive-2B despite utilizing similar parameters. Furthermore, \method-8B~outperforms VLM baselines, with its SFT and RL variants surpassing InternVL3-8B baseline by 3.4 and 4.1 PDMS, respectively. Notably, the improvements in No at-Fault Collision (NC) and Drivable Area Compliance (DAC) are attributed to the enhanced 3D spatial awareness, while the superior Time-to-Collision (TTC) and Ego Progress (EP) performance validates the dynamic foresight capabilities distilled from the world model.
Consistently, on the NAVSIMv2 benchmark (Table~\ref{table:navsimv2}), \method~maintains its superiority, achieving a new SOTA with 87.1 EPDMS. This score outperforms the previous best method, DriveVLA-W0-7B, by 1.0 EPDMS. Remarkably, even the lightweight 2B variant delivers exceptional performance with 86.8 EPDMS, which also surpasses the prior SOTA.
These results demonstrate that, compared to standard VLMs, our method significantly enhances comprehensive driving capabilities, particularly in complex scenarios that require 3D reasoning and future prediction. Moreover, the consistent excellence across both benchmarks confirms that our approach is not merely overfitting to the PDMS; rather, it demonstrates strong generalization by performing robustly on the more holistic and challenging EPDMS.

\noindent \textbf{SURDS and NuDynamics Benchmark.} 
Table~\ref{table:drivemllm} presents the comparative evaluation on both the SURDS and NuDynamics benchmarks. With only SFT, \method-8B and \method-2B demonstrate consistent superiority over their respective InternVL3 baselines. On the SURDS benchmark, \method-8B achieves improvements of 43.49\% and 7.68\% over the SURDS-3B method and InternVL3-8B baseline, respectively, while \method-2B delivers an 8.17\% improvement over InternVL3-2B. Specifically, \method-8B dominates intrinsic geometric reasoning, with Yaw Determination and Pixel Estimation reaching 70.16\% and 71.28\%, demonstrating precise absolute localization capabilities. Furthermore, exceptional performance on relational tasks, such as Left/Right (90.27\%) and Front/Behind (88.00\%), highlights its robust spatial compositional reasoning, mitigating the spatial disorientation often observed in prior VLMs. On NuDynamics, \method-8B and \method-2B show exceptional dynamic scene understanding with Motion scores of 81.19\% and 71.80\% respectively, outperforming both their fine-tuned baselines and the general Qwen2.5-VL-72B~\cite{bai2025qwen2}. This validates that aligning with world models endows the planner with the capability to accurately perceive and predict the motion states of dynamic agents across different parameter scales.
\begin{table}[t]
\centering
\small %
\setlength{\tabcolsep}{3.6pt}
\caption{Ablation study analyzing the impact of geometric (3D) and dynamic (WM) latent CoT across SFT and RL training phases.}
\begin{tabularx}{\linewidth}{l|cc|ccccc|c}
\toprule
Mode & WM & 3D & NC$\uparrow$ & DAC$\uparrow$ & TTC$\uparrow$ & CF$\uparrow$ & EP$\uparrow$ & PDMS$\uparrow$ \\
\midrule
SFT &  &  & 98.3 & 92.3 & 94.7 & 100 & 77.2 & 83.9 \\
 & \ding{51}  &  & 98.6 & 94.4 & 95.5 & 100 & 78.4 & 85.8 \\
 & & \ding{51} & 98.6 & 94.8 & \textbf{95.7} & 100 & 79.1 & 86.4 \\
 & \ding{51} & \ding{51} &\textbf{ 98.7} & \textbf{95.4} & \textbf{95.7} & \textbf{100} & \textbf{80.5} & \textbf{87.3} \\
\midrule
RL &  &  & 98.3 & 94.0 & 94.7 & 100 & 83.0 & 87.2  \\
 & \ding{51} &  & 98.5 & 97.7 & 95.2 & 100 & 86.7 & 90.9 \\
 &  & \ding{51} & 98.6 & 97.7 & 95.6 & 100 & 86.4 & 91.0 \\
 & \ding{51} & \ding{51} & \textbf{98.7} & \textbf{97.9} & \textbf{95.6} & \textbf{100} & \textbf{86.7} & \textbf{91.3} \\
\bottomrule
\end{tabularx}
\label{table:ablation1}
\vskip -2em
\end{table}

\subsection{Ablation Studies}

\noindent \textbf{On Latent Spatio-temporal Reasoning.}
Table~\ref{table:ablation1} validates the impact of our spatiotemporal reasoning components. Explicitly incorporating both geometric (3D) and dynamic (WM) pathways yields consistent gains across training phases. During SFT, the full model achieves a 3.4 PDMS increase over the baseline, driven by a substantial rise in DAC. This confirms that the geometric latent CoT grounds the planner in physical reality, ensuring precise adherence to boundaries. The RL phase further amplifies this advantage, widening the performance gap to a 4.1 PDMS improvement. Notably, the superior TTC and EP metrics highlight the role of the dynamic latent CoT: by distilling foresight from the World Model, the agent anticipates evolving dynamics for safer and more efficient planning.

\begin{table}[t]
\centering
\small
\setlength{\tabcolsep}{3.5pt}
\caption{Ablation study on the effectiveness of Textual CoT (Text), Latent CoT (Latent), and Latent Token Supervision(sup.) after RL training phase. ``w/o'' denotes ``without''.}
\begin{tabularx}{\linewidth}{l|ccccc|c}
\toprule
Method & NC$\uparrow$ & DAC$\uparrow$ & TTC$\uparrow$ & CF$\uparrow$ & EP$\uparrow$ & PDMS$\uparrow$ \\
\midrule
No Text  & 98.5 & 94.3 & 95.5 & 100 & 79.6 & 86.0 \\
Text  & 98.3 & 94.0 & 94.7 & 100 & 83.0 & 87.2 \\
Latent~(w/o sup.) & 98.6 & 96.8 & 95.8 & 100 & 84.6 & 89.8 \\
\midrule
\rowcolor{gray!30} Latent~(with sup.)  & \textbf{98.7} & \textbf{97.9} & \textbf{95.6} & \textbf{100} & \textbf{86.7} & \textbf{91.3} \\
\bottomrule
\end{tabularx}
\label{table:ablation2}
\vskip -1em
\end{table}

\begin{figure}[t]
    \centering
    \includegraphics[width=1.0\linewidth]{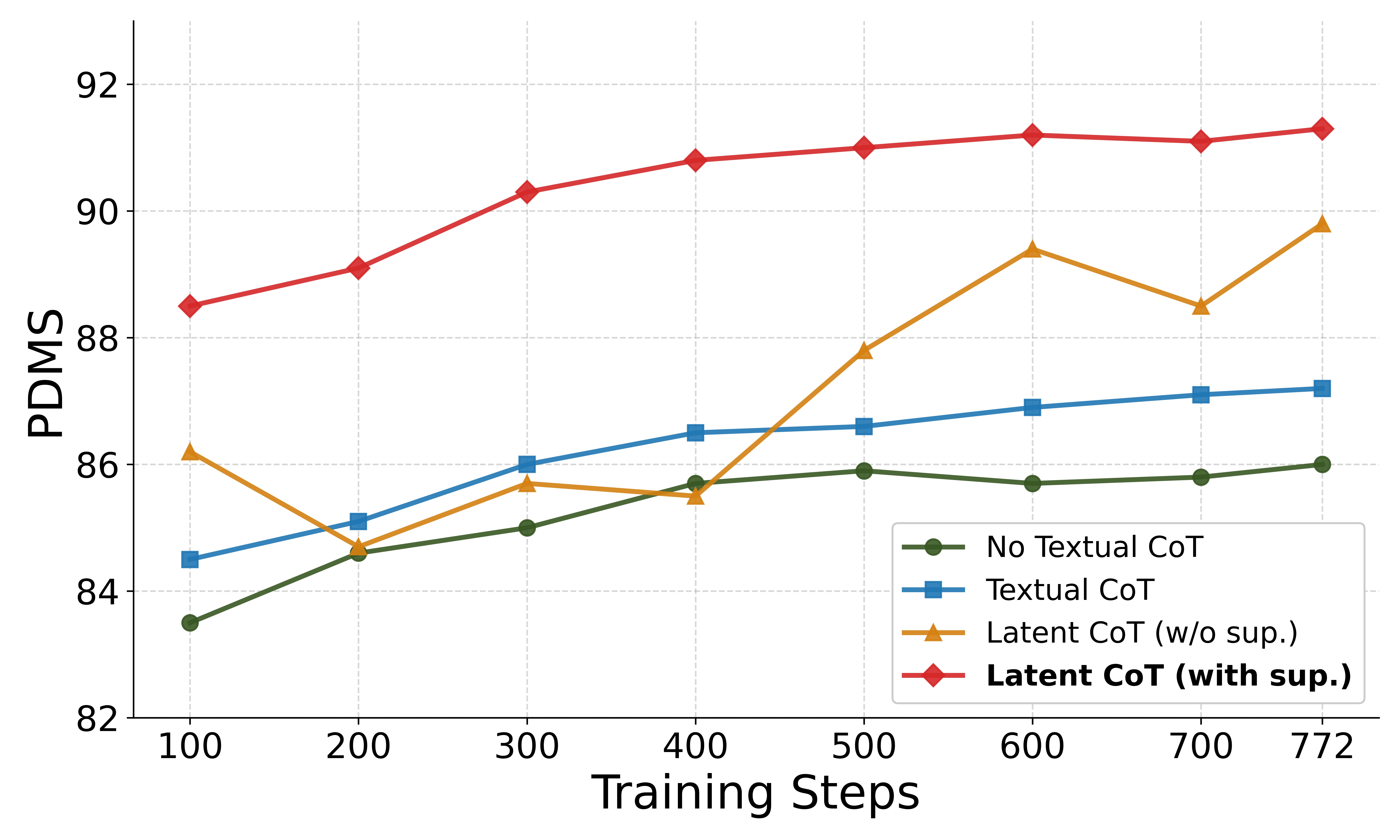}
    \caption{PDMS performance varying with training steps during the RL phase.}    
    \label{fig:pdms_step}
    \vskip -2.0em
\end{figure}

\begin{figure*}
    \centering
    \includegraphics[width=1.0\linewidth]{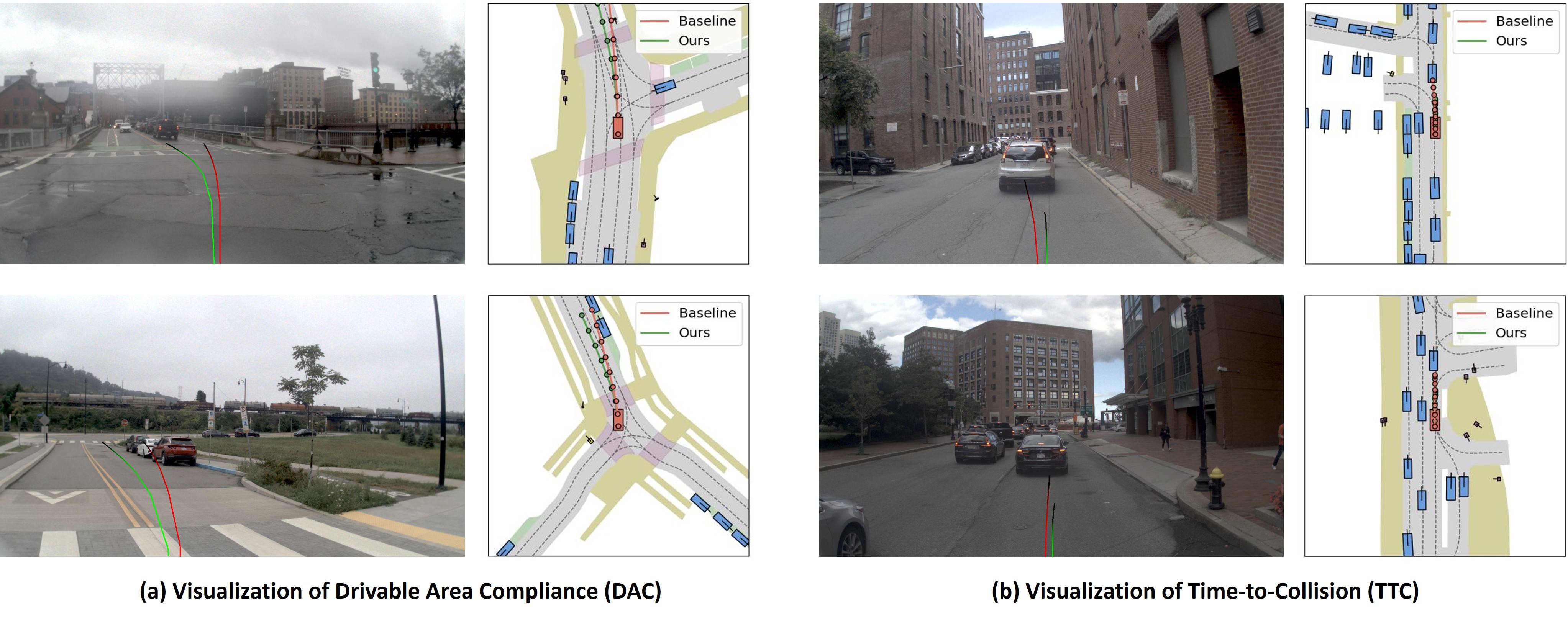}
    \vskip -0.5em
    \caption{Qualitative visualization comparing the \textbf{Textual CoT baseline (Red)} and \textbf{LaST-VLA (Green)}. (a) \textbf{Drivable Area Compliance (DAC):} Our method maintains precise lane adherence, whereas the baseline violates spatial boundaries. (b) \textbf{Time-to-Collision (TTC):} Our method accurately anticipates dynamics to avoid rear-end collisions, while the baseline fails to brake effectively.}
    \label{fig:visual}
\end{figure*}

\noindent \textbf{On Reasoning.}
Table~\ref{table:ablation2} investigates reasoning modalities. Compared to baseline without the textual CoT, introducing explicit textual CoT improves PDMS by 1.2, suggesting symbolic planning aids task decomposition but is limited by the semantic gap. Employing the latent CoT without supervision achieves 89.8 PDMS, effectively bypassing the textual bottleneck. However, this unsupervised approach functions as an ungrounded ``black box'' lacking physical constraints. This deficiency manifests as training instability: as visualized in Figure~\ref{fig:pdms_step}, the unsupervised variant suffers from marked performance oscillations even at later stages (orange line). In contrast, explicit physical supervision on latent tokens yields optimal performance, surpassing the unsupervised variant by 1.5 PDMS while achieving robust stability upon convergence (red line). Notably, this physical grounding translates into critical safety gains, boosting DAC to 97.9 and TTC to 95.6. These results demonstrate that aligning latent CoT with physical priors transforms unstable features into a robust, grounded reasoning engine.

\noindent \textbf{On Structured Causal Mask.}
Table~\ref{table:ablation_mask} examines the impact of the proposed structured causal mask. Compared to the standard masking baseline, our design increases PDMS by 2.0. This quantitative gain validates that blocking the direct visual shortcut effectively compels the planner to rely on the high-level latent spatio-temporal reasoning, thereby enhancing both trajectory compliance and safety.

\begin{table}[t]
\centering
\small %
\setlength{\tabcolsep}{2pt}
\caption{Ablation study on the effectiveness of the Structured Causal Mask after RL training phase.}
\begin{tabularx}{\linewidth}{l|ccccc|c}
\toprule
Mode  & NC$\uparrow$ & DAC$\uparrow$ & TTC$\uparrow$ & CF$\uparrow$ & EP$\uparrow$ & PDMS$\uparrow$ \\
\midrule
Standard Mask & 98.1 & 97.3 & 94.6 & 100 & 86.6 & 90.4 \\
\rowcolor{gray!30} Structured Causal Mask & \textbf{98.7} & \textbf{97.9} & \textbf{95.6} & \textbf{100} & \textbf{86.7} & \textbf{91.3} \\
\bottomrule
\end{tabularx}
\vskip -1.5em
\label{table:ablation_mask}
\end{table}

\noindent \textbf{On the Number of Latent Tokens.}
Table~\ref{table:ablation4} examines the impact of latent granularity. The configuration with $N_{\text{3D}}=12$ and $N_{\text{WM}}=3 \times 12$ yields the optimal 91.3 PDMS. Reducing tokens degrades performance due to information bottlenecks, while increasing them introduces redundancy that complicates optimization. Thus, the selected configuration balances capacity and efficiency. Overall, the model performance remains relatively stable across varying token numbers.

\subsection{Qualitative Results}
To visualize the efficacy of our approach, Figure~\ref{fig:visual} presents a qualitative comparison between the textual CoT baseline and LaST-VLA. As illustrated in Figure~\ref{fig:visual}(a), LaST-VLA (Green) generates smooth, compliant trajectories that strictly adhere to lane boundaries, whereas the baseline (Red) struggles with metric precision, occasionally deviating from the drivable area. This validates that our geometric latent CoT, explicitly aligned with 3D foundation models, effectively grounds the planner with accurate spatial constraints. Furthermore, regarding dynamic safety in Figure~\ref{fig:visual}(b), while the baseline suffers from limited foresight and fails to decelerate, LaST-VLA accurately predicts future dynamics and executes timely braking to ensure safety in critical interaction scenarios. This demonstrates the significance of the dynamic latent CoT in distilling temporal foresight from the world model, equipping the planner with predictive capabilities typically lacking in pure textual reasoning. More results can be found in Appendix~\ref{appendix:exp_results}.

\begin{table}[t]
\centering
\small 
\setlength{\tabcolsep}{3.2pt}
\caption{Ablation study on the number of geometric(3D) and dynamic(WM) latent tokens after RL training phase.}\begin{tabularx}{\linewidth}{l|cc|ccccc|c}
\toprule
ID & WM/N & 3D/N & NC$\uparrow$ & DAC$\uparrow$ & TTC$\uparrow$ & CF$\uparrow$ & EP$\uparrow$ & PDMS$\uparrow$ \\
\midrule
1 & 3*12 & 6 & 98.6 & 97.5 & 95.6 & 100 & 86.5 & 90.9 \\
2 & 3*12 & 24 & 98.4 & 97.7 & 95.2 & 100 & 86.5 & 90.8 \\
3 & 3*6 & 12 & 98.6 & 97.5 & 95.4 & 100 & 86.5 & 90.8 \\
4 & 3*24 & 12 & \textbf{98.9} & 97.6 & \textbf{96.0} & 100 & 85.9 & 90.9 \\
\midrule
\rowcolor{gray!30}  5 & 3*12 & 12 & 98.7 & \textbf{97.9} & 95.6 & \textbf{100} & \textbf{86.7} & \textbf{91.3} \\
\bottomrule
\end{tabularx}
\label{table:ablation4}
\vskip -1.5em
\end{table}

%% file: sec/5_conclusion.tex
\section{Conclusion}
In this work, we introduced \textbf{\method}, a framework that shifts the reasoning paradigm of autonomous driving from explicit text to a continuous Latent Spatio-Temporal Space. By distilling physical priors from 3D and video foundation models, our approach effectively addresses the challenges of inference latency, semantic hallucinations, and the lack of physical grounding inherent to textual CoT. Coupled with a progressive training strategy and GRPO, \method~achieves SOTA performance on NAVSIM, SURDS and NuDynamics benchmarks. These results demonstrate that aligning latent thinking with physical reality significantly enhances the robustness, efficiency, and safety of VLA-based planning.

%% file: sec/appendix.tex
\section{Appendix} \label{appendix:appendix} In this appendix, we present details to support the findings of the main paper. First, \textbf{Section~\ref{appendix:exp_results}} reports extended experimental results, including open-loop evaluations on the nuScenes benchmark, comprehensive ablation studies regarding geometric supervision. \textbf{Section~\ref{appendix:data_prepare}} elaborates on the data preparation protocols for the nuScenes and NuDynamics benchmarks. \textbf{Section~\ref{appendix:exp_details}} outlines the implementation specifications, covering model architecture, training parameters, evaluation metrics, and RL reward design. Finally, \textbf{Section~\ref{appendix:vis}} provides qualitative visualizations of successful and failure planning scenarios, along with supplementary assessments on the SURDS and NuDynamics benchmarks.

\section{More Results} \label{appendix:exp_results}
\noindent{\textbf{nuScenes Benchmark.}}\footnote{This work masks use of the nuScenes dataset (https://www.nuscenes.org/). The authors of this work confirm that the use of the above dataset in this work is strictly limited to academic research purposes and does not involve any commercial activities.}
Table~\ref{tab:nuscenes} presents the open-loop planning evaluation on the nuScenes dataset. \method-8B achieves SOTA performance in trajectory precision, recording the lowest average L2 error of 0.38m among all compared VLM-based methods. Beyond superior accuracy, the model maintains an low collision rate of 0.18\%, demonstrating that our latent reasoning paradigm effectively balances precise geometric adherence with safety. Furthermore, the lightweight \method-2B also delivers robust performance, validating the effectiveness of our approach across different model sizes.

\begin{table}[h]
\centering
\caption{Comparison with state-of-the-art methods on the nuScenes. * indicates that the ego status is use.}
\begin{tabular}{l|l|cccc|cccc} 
\toprule
\multirow{2}{*}{Method} & \multirow{2}{*}{VLM Model} & \multicolumn{4}{c|}{L2 (m) $\downarrow$} & \multicolumn{4}{c}{Collision (\%) $\downarrow$} \\
\cmidrule(lr){3-6} \cmidrule(lr){7-10} 
 & & 1s & 2s & 3s & Avg. & 1s & 2s & 3s & Avg. \\
\midrule
\multicolumn{10}{l}{\textbf{\emph{Conventional End-to-end Methods}}} \\
UniAD*~\cite{hu2023planning} & - & 0.20 & 0.42 & 0.75 &0.46& 0.02& 0.25& 0.84& 0.37 \\
\midrule
\multicolumn{10}{l}{\textbf{\emph{VLMs-based Methods}}} \\
Doe-1~\cite{zheng2024doe} & Lumina-mGPT-7B & 0.50 & 1.18 & 2.11 & 1.26 & 0.04 & 0.37 & 1.19 & 0.53 \\
AutoVLA*~\cite{zhou2025autovla} & Qwen2.5-VL-3B & 0.28 & 0.66 & 1.16 & 0.70 & 0.14 & 0.25 & 0.53 & 0.31 \\
RDA-Driver*~\cite{huang2024making} & LLaVA-7B & 0.23 & 0.73 & 1.54 & 0.80 & 0.00 & 0.13 & 0.83 & 0.32 \\
OpenDriveVLA*~\cite{zhou2025opendrivevla} &  Qwen2.5VL-3B & 0.19 & 0.58 & 1.24 & 0.67 & 0.02 & 0.18 & 0.70 & 0.30 \\
FSDrive*~\cite{zeng2025futuresightdrive} & Qwen2-VL-2B & 0.18& 0.39& 0.77& 0.45& \textbf{0.00}& \textbf{0.06} & \textbf{0.42} & \textbf{0.16} \\
\midrule
\rowcolor{gray!30} \textbf{\method*~(Ours)} & InternVL3-2B & 0.19 & 0.35 & 0.71 & 0.42 & 0.03 & 0.16 & 0.47 & 0.22 \\
\rowcolor{gray!30} \textbf{\method*~(Ours)} & InternVL3-8B & \textbf{0.17} & \textbf{0.33} & \textbf{0.64} & \textbf{0.38} & \textbf{0.00} & 0.11 & \textbf{0.42} & 0.18 \\
\bottomrule
\end{tabular}
\label{tab:nuscenes}
\end{table}

\noindent{\textbf{Ablation on Geometric Supervision Layers.}}
Table~\ref{table:ablation_features} investigates the efficacy of supervising latent thoughts with different layers of the VGGT aggregator. We observe that supervising shallow layers (Features 4 and 11) yields suboptimal performance. This is likely because these early layers primarily encode low-level visual textures rather than the explicit 3D depth and geometric structures essential for spatial planning. While performance improves with increasing depth (Feature 17), aggregating features across all layers (ID 4) fails to outperform the single final layer. We attribute this diminishing return to increased learning difficulty due to information redundancy and the need for a longer latent CoT sequence, which complicates optimization. In contrast, supervising solely with the final layer (Feature 23) achieves the peak performance of 91.0 PDMS, confirming that the terminal layer encapsulates the most refined and compact 3D geometric priors for efficient reasoning.

\begin{table}[h]
\centering
\setlength{\tabcolsep}{6pt} 
\caption{Ablation study on the effectiveness of supervising features from different layers of the VGGT aggregator. The Feature~(23) is the last layer. \textbf{Note: These methods only have geometric (3D) latent CoT, without dynamic (WM) latent CoT.}}
\begin{tabular}{c|p{3.5cm}<{\centering}|ccccc|c}
\toprule
ID & Supervised Features & NC$\uparrow$ & DAC$\uparrow$ & TTC$\uparrow$ & CF$\uparrow$ & EP$\uparrow$ & PDMS$\uparrow$ \\
\midrule
1 & Feature~(4) & 98.3 & 97.3 & 94.8 & \textbf{100} & 86.5 & 90.4 \\
2 & Feature~(11) & 98.2 & 97.3 & 94.6 & \textbf{100} & 86.5 & 90.3 \\
3 & Feature~(17) & 98.4 & 97.5 & 95.3 & \textbf{100} & 86.4 & 90.7 \\
4 & Feature~(4, 11, 17, 23) & 98.4 & \textbf{97.7} & 94.8 & \textbf{100} & \textbf{86.8} & 90.8 \\
\midrule
\rowcolor{gray!30} 5 & \textbf{Feature~(23)~(Ours)} & \textbf{98.6} & \textbf{97.7} & \textbf{95.6} & \textbf{100} & 86.4 & \textbf{91.0} \\
\bottomrule
\end{tabular}
\label{table:ablation_features}
\end{table}

\section{Data Preparation Details} \label{appendix:data_prepare}
\subsection{nuScenes}
To further validate the generalization and effectiveness of our proposed Spatio-Temporal Latent CoT, we extend our evaluation to the \textbf{nuScenes} dataset~\cite{caesar2020nuscenes}. This large-scale benchmark comprises 1,000 complex driving scenarios, each spanning approximately 20 seconds. The data is captured via a comprehensive sensor suite, featuring a 32-beam LiDAR and six surrounding cameras that provide a 360-degree field of view. For our experiments, we utilize the standard split containing 28,130 samples for training and 6,019 samples for validation.

\subsection{NuDynamics} \label{appendix:nudynamics}
NuDynamics is constructed upon the SURDS dataset with the goal of evaluating dynamic scene understanding capabilities. We select clearly visible targets from the NuScenes dataset and utilize Qwen2.5-VL-32B ~\cite{bai2025qwen2} to generate concise descriptions. Behavioral labels, including stopped crossing, moving in the same direction, moving in the opposite direction, and diagonal movement, are derived by mapping native annotations from the nuScenes dataset. The dataset comprises 4K training samples and 1K evaluation samples, and a prompt example is illustrated in Figure~\ref{fig:nudynamics}.

\section{Experimental Details}  \label{appendix:exp_details}
\noindent{\textbf{Model Architecture.}}
We leverage InternVL3-8B~\cite{zhu2025internvl3} as our vision-language backbone, bridging a 300M-parameter InternViT encoder with the Qwen2.5-7B language model. Distinguished by its resolution-adaptive architecture, the model dynamically modulates feature extraction granularity according to image complexity. By prioritizing fine-grained processing for information-rich regions while applying coarser abstraction to simpler areas, it effectively strikes an optimal balance between visual fidelity and computational overhead.

\noindent{\textbf{Training Parameters and Hardware Configuration.}} \label{appendix:param}
We utilize InternVL3-8B~\cite{zhu2025internvl3} as the foundation model, training across three sequential stages on 8 NVIDIA H20 GPUs. The first SFT stage fine-tunes on the navtrain-hard-24k subset for 2 epochs with a learning rate of $1 \times 10^{-5}$, batch size of 2, and 4 gradient accumulation steps; we enable the Structured Causal Mask and set loss weights to $\lambda_{WM}=\lambda_{3D}=1.00$ and $\lambda_{action}=0.01$. Subsequently, the second stage trains on the full 85k navtrain dataset for 2 epochs using 2 gradient accumulation steps, where the mask is disabled and weights are adjusted to $\lambda_{action}=1.00$ with $\lambda_{WM}=\lambda_{3D}=0.01$. Finally, the RL phase employs GRPO for 2 iterations using a learning rate of $2 \times 10^{-6}$, batch size of 2, 16 gradient accumulation steps, 8 generations, and a sampling temperature of 2.0, with reward coefficients set to $\lambda_{traj}=8, \lambda_{fmt}=1, \lambda_{goal}=1$ and a KL penalty $\beta=0.1$. The training duration for the three stages is approximately 1 hour, 3.5 hours, and 48 hours, respectively.

\noindent{\textbf{Metric.}} \label{appendix:metric}
To comprehensively evaluate planning performance, we employ specific metrics tailored to each benchmark. For closed-loop planning within the NAVSIM suite, we adopt the Predictive Driver Model Score (PDMS) for \textbf{NAVSIMv1}~\cite{dauner2024navsim} and the Extended Predictive Driver Model Score (EPDMS) for \textbf{NAVSIMv2}~\cite{cao2025pseudo}. Additionally, on the nuScenes dataset, we follow the protocol of UniAD~\cite{hu2023planning} to assess open-loop trajectory quality using L2 metrics and collision rate.

For NAVSIMv1, PDMS integrates five sub-metrics: No At-Fault Collision (NC), Drivable Area Compliance (DAC), Time-to-Collision (TTC), Comfort (C), and Ego Progress (EP) to produce a comprehensive closed-loop planning score. Its calculation formula is defined as follows:
\begin{equation}
PDMS = NC \times DAC \times \left( \frac{5\times EP + 5\times TTC + 2\times C}{12} \right),
\end{equation}

For NAVSIMv2, the EPDMS metric includes several components categorized as penalties or weighted subscores. Its key metrics are No at-Fault Collision (NC), Drivable Area Compliance (DAC), Driving Direction Compliance (DDC), Traffic Light Compliance (TLC), Ego Progress (EP), Time to Collision (TTC), Lane Keeping (LK), History Comfort (HC), and Extended Comfort (EC). Its calculation formula is defined as follows:
\begin{equation}
\begin{split}
    EPDMS ={} & NC \times DAC \times DDC \times TLC \times \left( \frac{5\times EP + 2\times LK + 2\times HC + 5\times TTC + 2\times EC}{16} \right).
\end{split}
\end{equation}

\noindent{\textbf{Detailed Rewards Design in RL.}} \label{appendix:reward}
To effectively guide the policy optimization during the Reinforcement Learning stage, we formulate a composite reward function comprising three distinct components: trajectory quality ($R_{\text{traj}}$), structural validity ($R_{\text{fmt}}$), and goal alignment ($R_{\text{goal}}$).

For the trajectory reward $r_{\text{traj}}$, we directly leverage the feedback from the NAVSIM simulator. Instead of using a surrogate loss, we utilize the continuous PDMS, ranging from 0 to 1, as the primary reward signal. This encourages the model to generate paths that maximize safety and drivability metrics as defined above.

For the \textbf{format reward} $r_{\text{fmt}}$, we assign a total of 1.0 point to enforce structural integrity. This is equally distributed: 0.5 points are awarded for the correct usage of structural tags ($<$latent\_start\_wm$>$...$<$latent\_end\_wm$>$, $<$latent\_start\_3d$>$...$<$latent\_end\_3d$>$, and $<$answer$>$...$<$/answer$>$), and the remaining 0.5 points validate the syntax of the trajectory waypoints to ensure the output is well-formed and machine-parsable.

For the \textbf{goal reward} $r_{\text{goal}}$, we employ a piecewise function based on the L1 distance to encourage precise alignment between the predicted endpoint and the ground truth~\cite{adathinkdrive}.

\section{Visualization Analysis} \label{appendix:vis}

\begin{figure*}[h]
    \centering
    \includegraphics[width=1.0\linewidth]{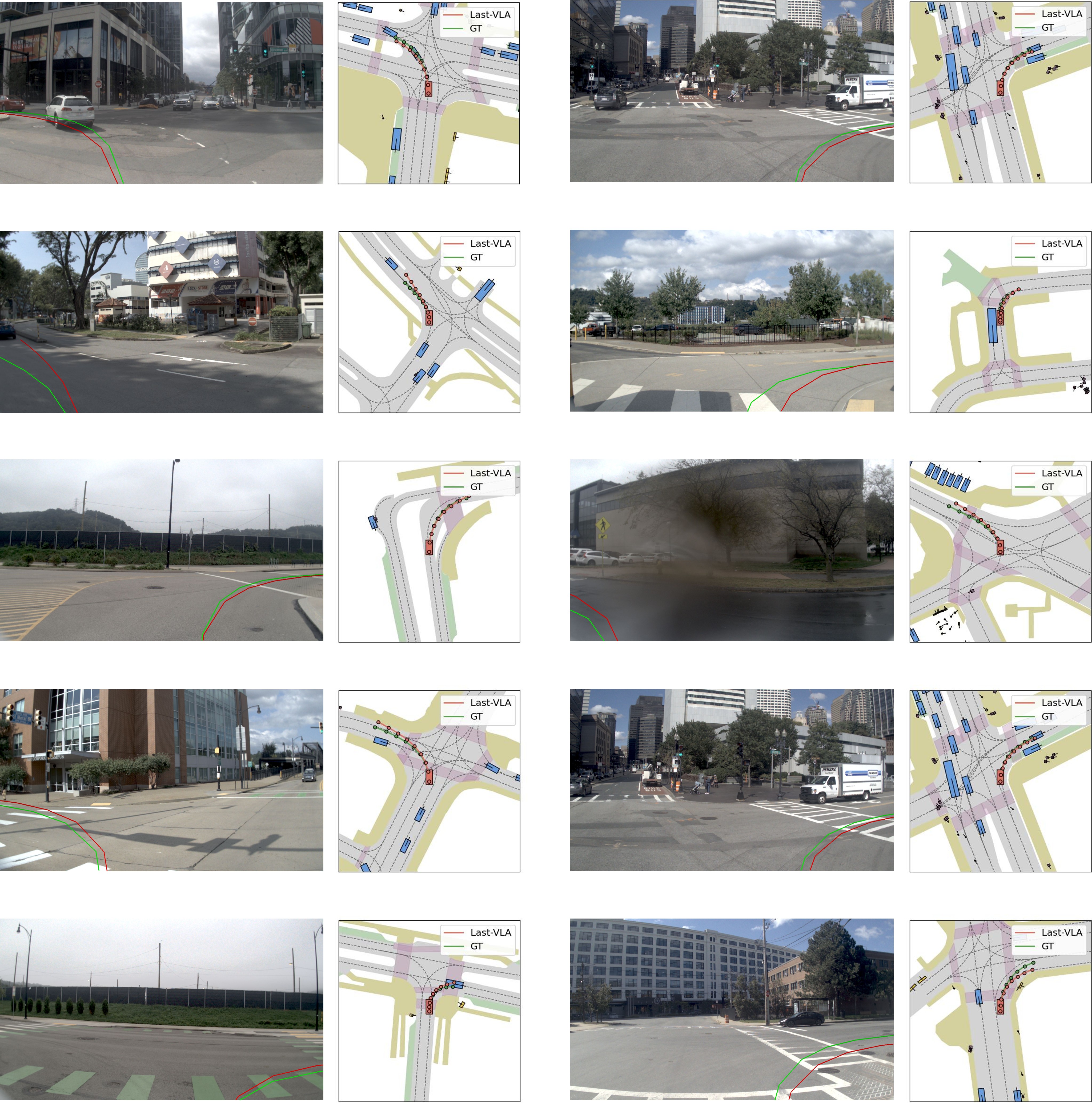}
    \caption{\textbf{Qualitative visualization of failure cases on NAVSIM benchmark.} The red line represents the trajectory predicted by \method, while the green line denotes the Ground Truth. We observe that failures predominantly occur when the planned path extends beyond the front-view camera's field of view (FOV). Lacking sufficient visual context in these peripheral regions, the model struggles with precise spatial grounding, leading to potential collisions or deviations from the drivable area.}    
    \label{fig:visual_failure}
\end{figure*}

\noindent\textbf{Failure Analysis and Limitations.}
We analyzed failure cases where the planning performance drops significantly, such as when the PDMS is 0. As shown in Figure~\ref{fig:visual_failure}, the main failure pattern occurs when the target trajectory extends beyond the Field of View (FOV) of the front-view camera. This typically happens in scenarios involving sharp turns or complex intersections, where the model lacks sufficient visual information about the peripheral environment. Consequently, the planner may generate paths that deviate from the drivable area. This issue is primarily due to our current reliance on a single front-view camera. In future work, we plan to address this by incorporating surround-view inputs and temporal information to expand the perceptual range for more robust planning.

\begin{figure*}[h]
    \centering
    \includegraphics[width=1.0\linewidth]{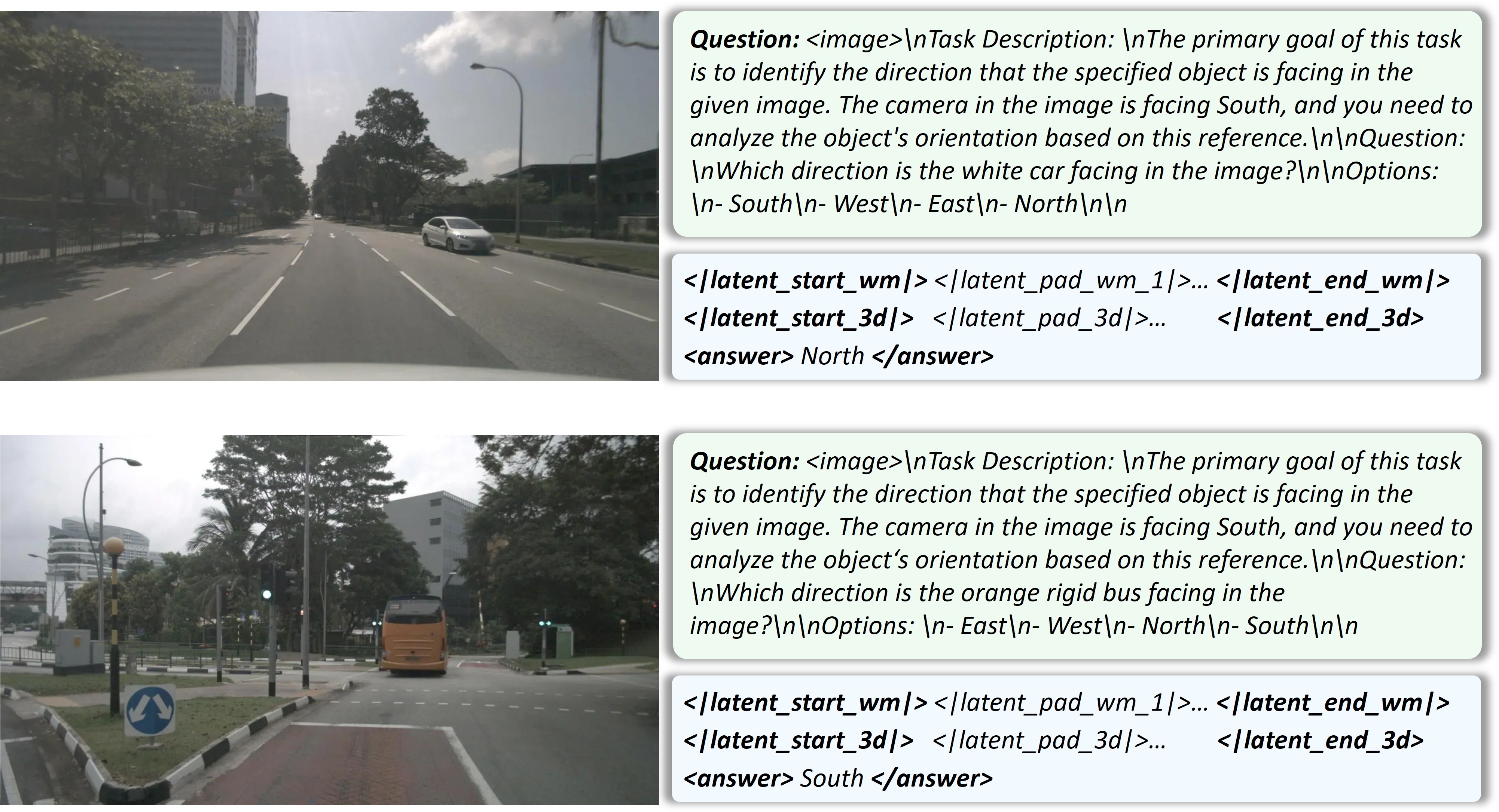}
    \caption{Examples of the Yaw Angle Determination task on the SURDS benchmark.}    
    \label{fig:surds1}
\end{figure*}

\begin{figure*}[h]
    \centering
    \includegraphics[width=1.0\linewidth]{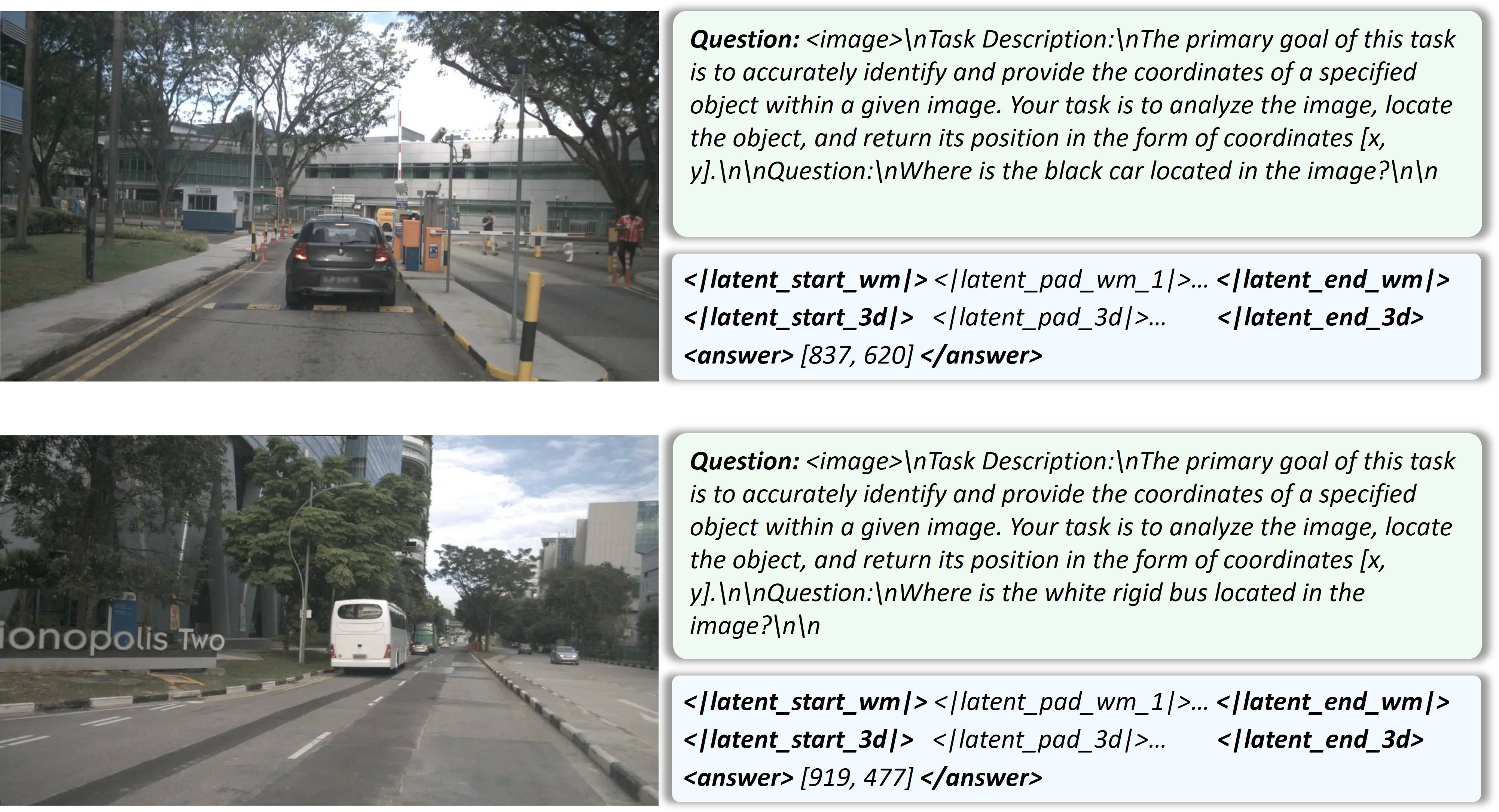}
    \caption{Examples of the Pixel Location Estimation task on the SURDS benchmark.}    
    \label{fig:surds2}
\end{figure*}

\begin{figure*}[h]
    \centering
    \includegraphics[width=1.0\linewidth]{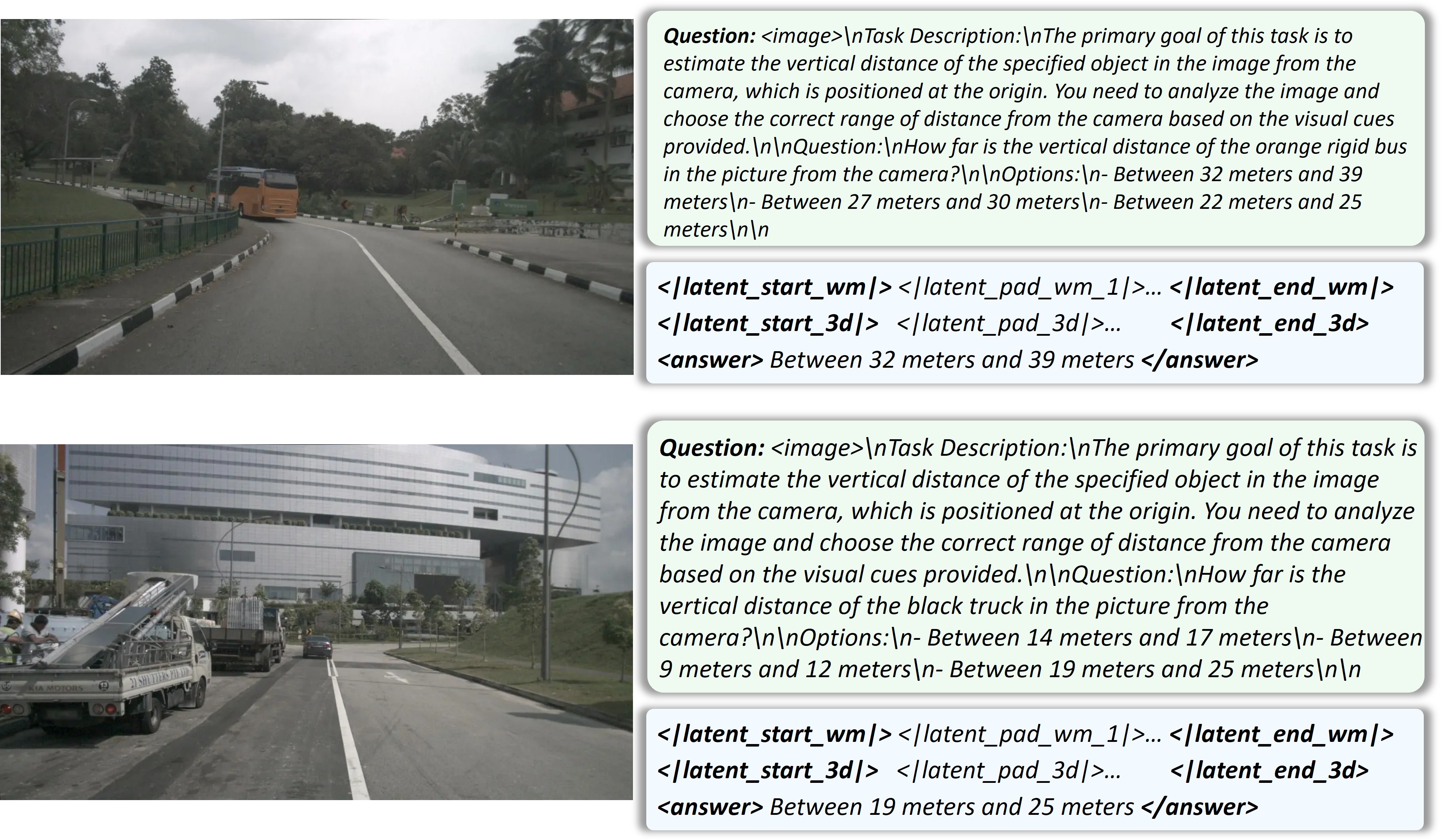}
    \caption{Examples of the Depth Range Determination on the SURDS benchmark.}    
    \label{fig:surds3}
\end{figure*}

\begin{figure*}[h]
    \centering
    \includegraphics[width=1.0\linewidth]{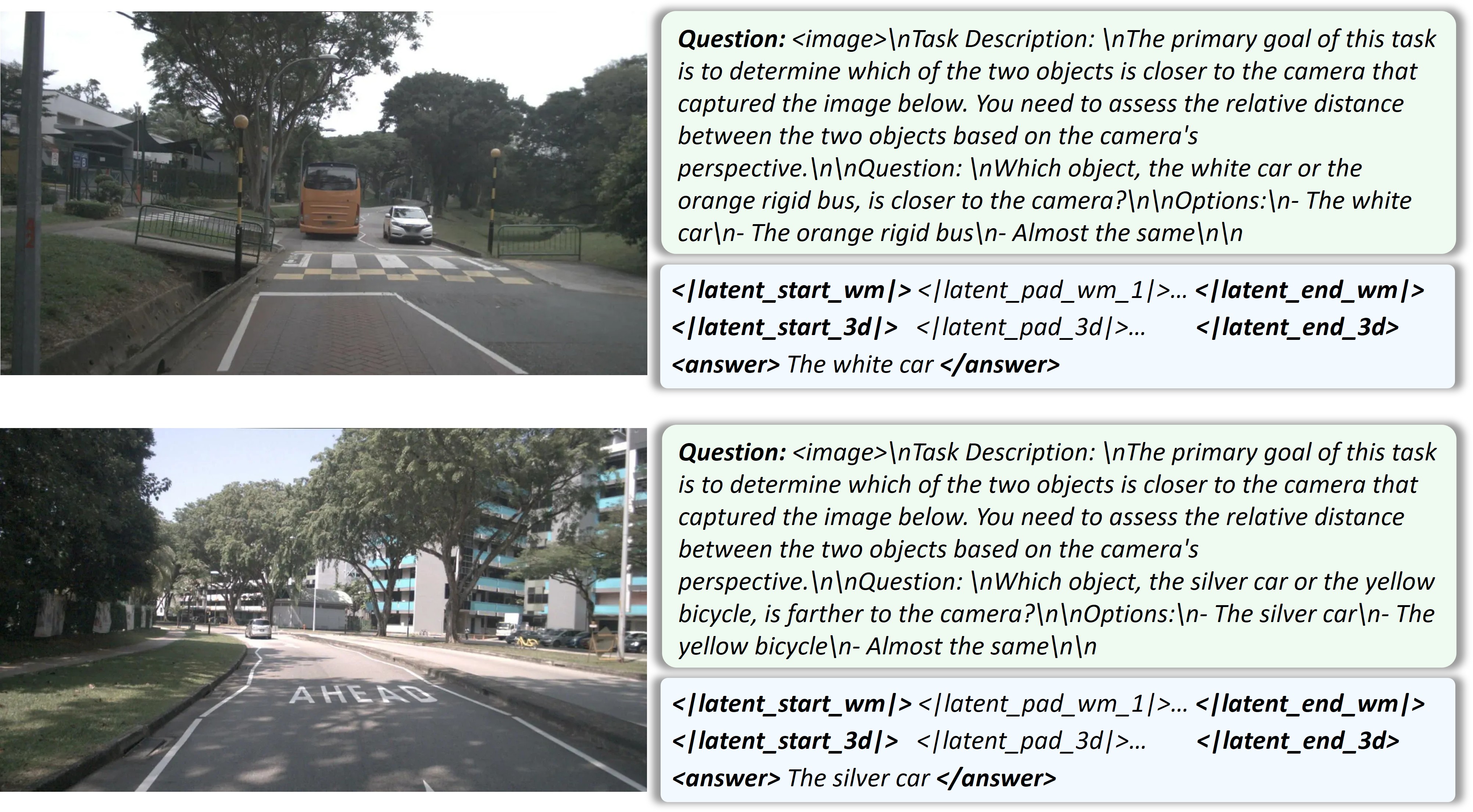}
    \caption{Examples of the Distance Estimation task on the SURDS benchmark.}    
    \label{fig:surds4}
\end{figure*}

\begin{figure*}[h]
    \centering
    \includegraphics[width=1.0\linewidth]{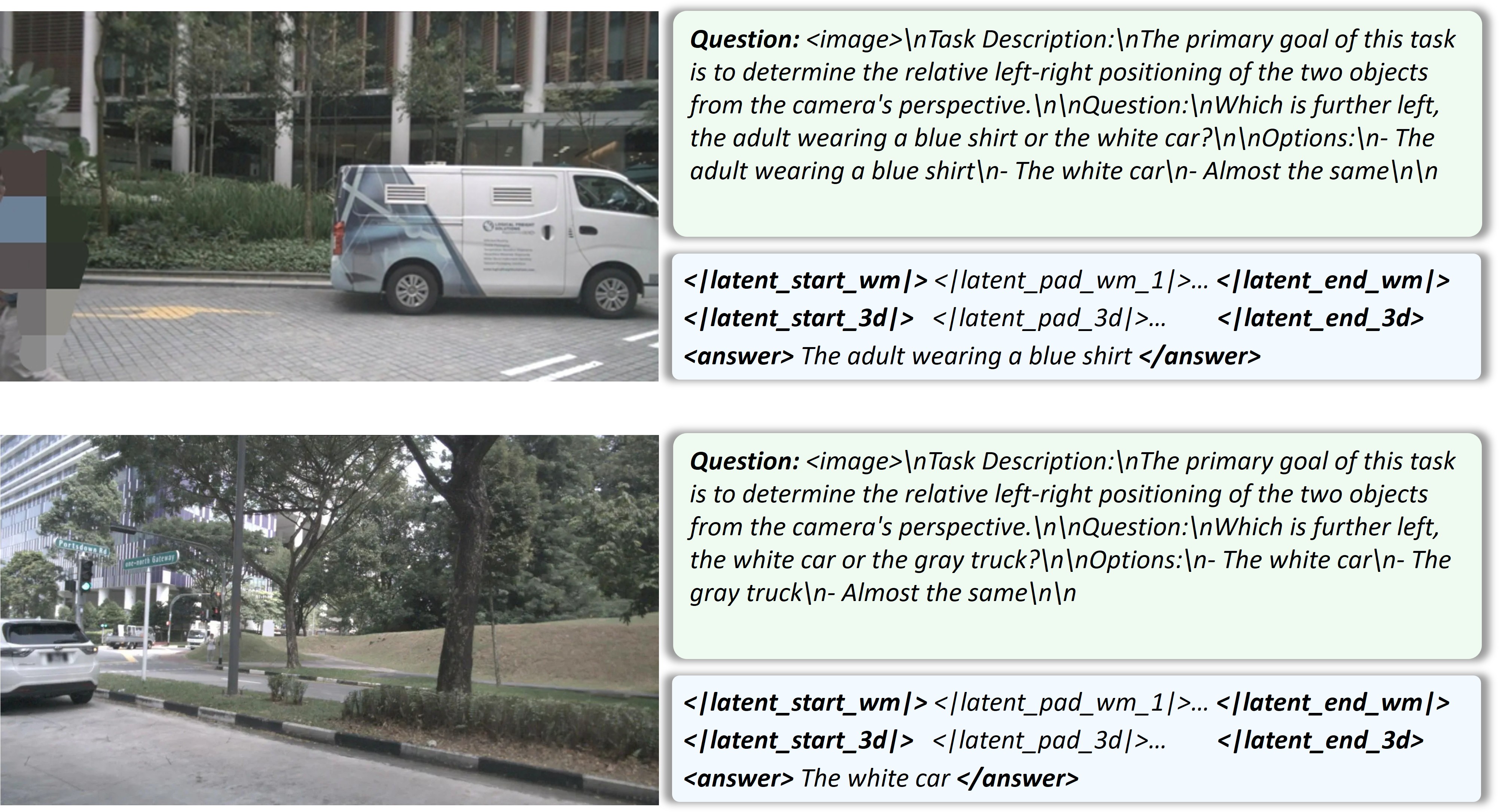}
    \caption{Examples of the Left/Right Determination task on the SURDS benchmark.}    
    \label{fig:surds5}
\end{figure*}

\begin{figure*}[h]
    \centering
    \includegraphics[width=1.0\linewidth]{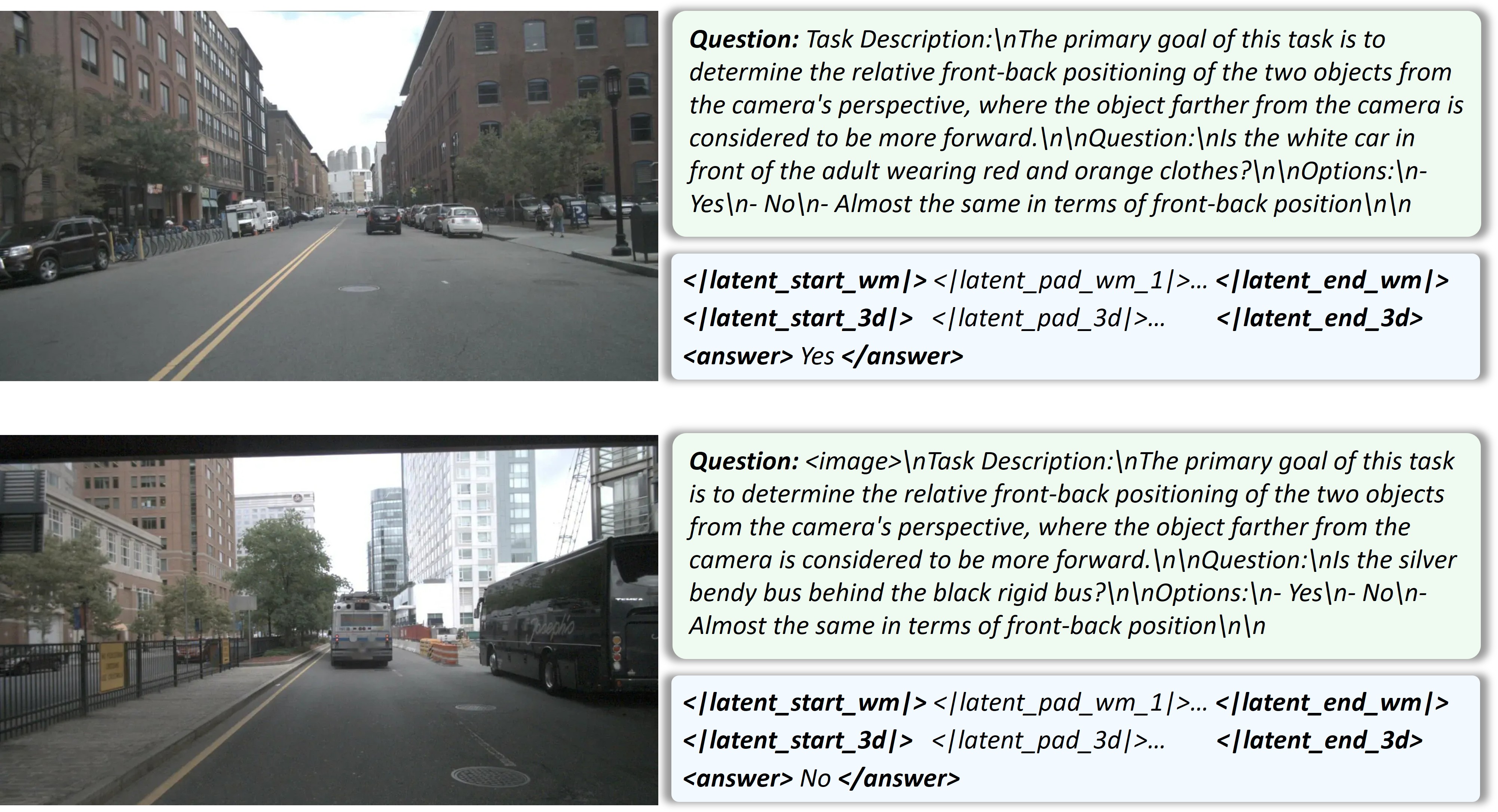}
    \caption{Examples of the Front/Behind Determination task on the SURDS benchmark.}    
    \label{fig:surds6}
\end{figure*}

\begin{figure*}[h]
    \centering
    \includegraphics[width=1.0\linewidth]{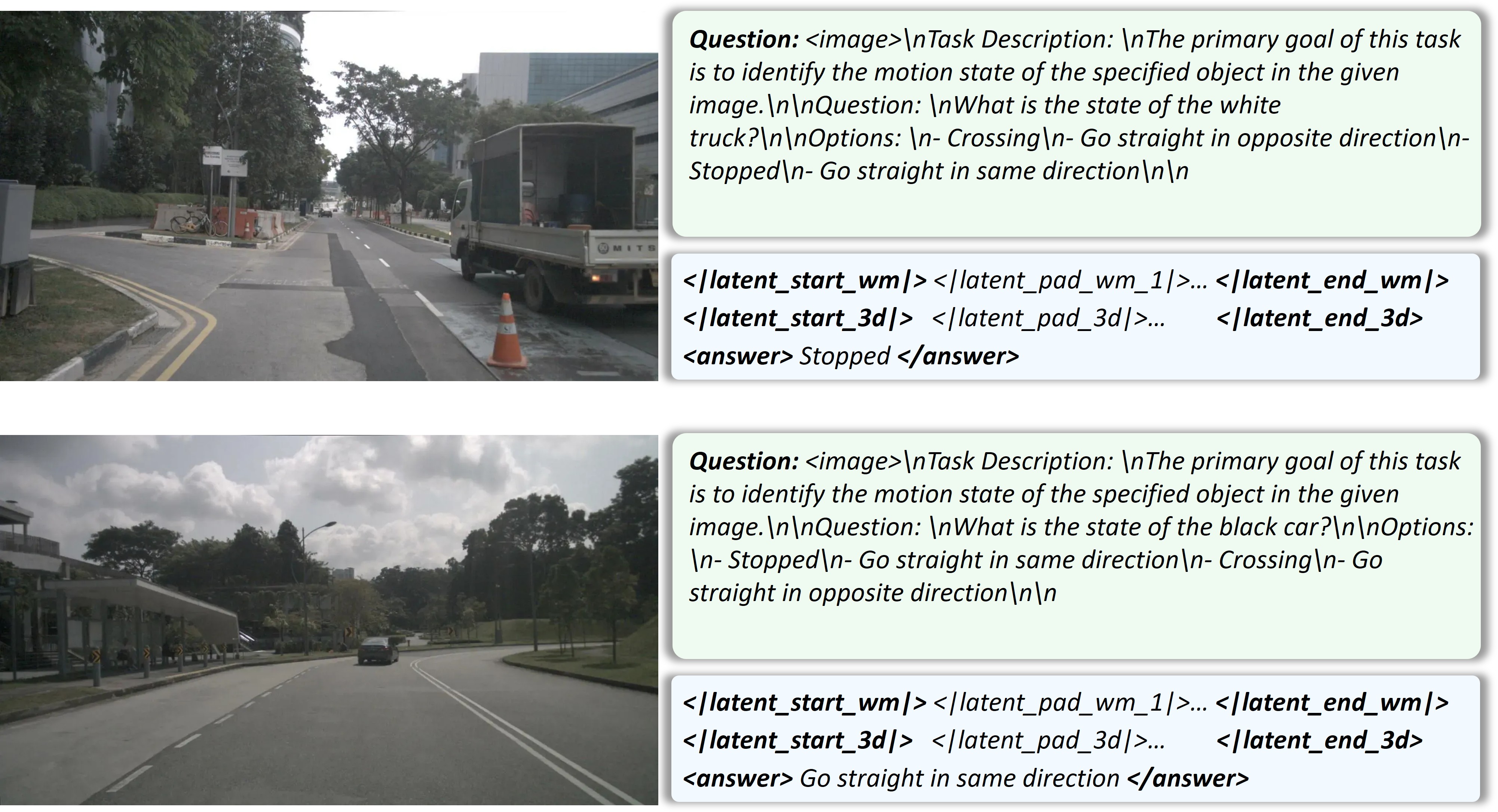}
    \caption{Examples of the  Motion State Estimation task on the NuDynamics benchmark.}    
    \label{fig:nudynamics}
\end{figure*}

\noindent\textbf{Qualitative Visualization on SURDS and NuDynamics.}
To further evaluate the comprehensive reasoning capabilities of \method, we present qualitative visualizations on the SURDS and NuDynamics benchmarks. Figures~\ref{fig:surds1} through \ref{fig:surds6} illustrate representative input queries and corresponding model responses across diverse SURDS tasks, such as Yaw Angle Determination and Pixel Location Estimation. These examples confirm that our model can accurately interpret complex 3D geometric constraints from 2D images. Furthermore, the visualizations on the NuDynamics benchmark (Figure~\ref{fig:nudynamics}) highlight the model's proficiency in analyzing scene evolution and motion trends. Collectively, these results demonstrate that our latent reasoning paradigm effectively equips the agent with robust 3D spatial awareness and dynamic scene understanding.
 

\begin{figure*}[h]
    \centering
    \includegraphics[width=1.0\linewidth]{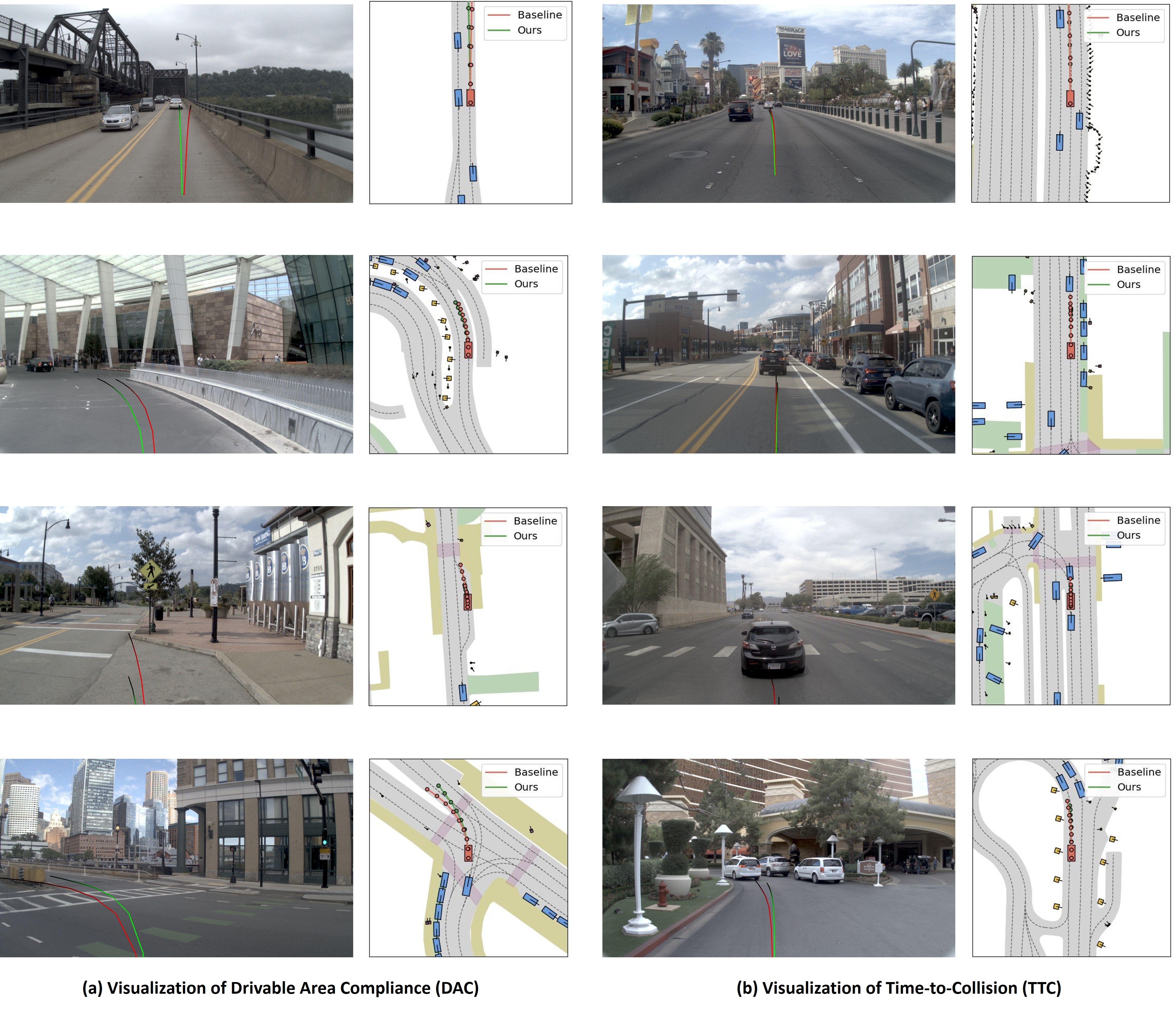}
    \caption{Qualitative visualization comparing the \textbf{Textual CoT baseline (Red)} and \textbf{LaST-VLA (Green)} on NAVSIM benchmark. (a) \textbf{Drivable Area Compliance (DAC):} Our method maintains precise lane adherence, whereas the baseline violates spatial boundaries. (b) \textbf{Time-to-Collision (TTC):} Our method accurately anticipates dynamics to avoid rear-end collisions, while the baseline fails to brake effectively.}
    \label{fig:visual_appendix}
\end{figure*}




%% file: example_paper.bib
@article{adathinkdrive,
  title={AdaThinkDrive: Adaptive Thinking via Reinforcement Learning for Autonomous Driving},
  author={Luo, Yuechen and Li, Fang and Xu, Shaoqing and Lai, Zhiyi and Yang, Lei and Chen, Qimao and Luo, Ziang and Xie, Zixun and Jiang, Shengyin and Liu, Jiaxin and others},
  journal={arXiv preprint arXiv:2509.13769},
  year={2025}
}

@article{team2023gemini,
  title={Gemini: a family of highly capable multimodal models},
  author={Team, Gemini and Anil, Rohan and Borgeaud, Sebastian and Alayrac, Jean-Baptiste and Yu, Jiahui and Soricut, Radu and Schalkwyk, Johan and Dai, Andrew M and Hauth, Anja and Millican, Katie and others},
  journal={arXiv preprint arXiv:2312.11805},
  year={2023}
}

@article{li2025latent,
  title={Latent visual reasoning},
  author={Li, Bangzheng and Sun, Ximeng and Liu, Jiang and Wang, Ze and Wu, Jialian and Yu, Xiaodong and Chen, Hao and Barsoum, Emad and Chen, Muhao and Liu, Zicheng},
  journal={arXiv preprint arXiv:2509.24251},
  year={2025}
}

@article{li2025drive,
  title={Drive-R1: Bridging Reasoning and Planning in VLMs for Autonomous Driving with Reinforcement Learning},
  author={Li, Yue and Tian, Meng and Zhu, Dechang and Zhu, Jiangtong and Lin, Zhenyu and Xiong, Zhiwei and Zhao, Xinhai},
  journal={arXiv preprint arXiv:2506.18234},
  year={2025}
}

@article{wei2025sim,
  title={SIM-CoT: Supervised Implicit Chain-of-Thought},
  author={Wei, Xilin and Liu, Xiaoran and Zang, Yuhang and Dong, Xiaoyi and Cao, Yuhang and Wang, Jiaqi and Qiu, Xipeng and Lin, Dahua},
  journal={arXiv preprint arXiv:2509.20317},
  year={2025}
}

@article{zhang2024sparsead,
  title={Sparsead: Sparse query-centric paradigm for efficient end-to-end autonomous driving},
  author={Zhang, Diankun and Wang, Guoan and Zhu, Runwen and Zhao, Jianbo and Chen, Xiwu and Zhang, Siyu and Gong, Jiahao and Zhou, Qibin and Zhang, Wenyuan and Wang, Ningzi and others},
  journal={arXiv preprint arXiv:2404.06892},
  year={2024}
}

@inproceedings{wang2023exploring,
  title={Exploring object-centric temporal modeling for efficient multi-view 3d object detection},
  author={Wang, Shihao and Liu, Yingfei and Wang, Tiancai and Li, Ying and Zhang, Xiangyu},
  booktitle={Proceedings of the IEEE/CVF international conference on computer vision},
  pages={3621--3631},
  year={2023}
}

@article{cheng2024compressed,
  title={Compressed chain of thought: Efficient reasoning through dense representations},
  author={Cheng, Jeffrey and Van Durme, Benjamin},
  journal={arXiv preprint arXiv:2412.13171},
  year={2024}
}

@article{zheng2025learning,
  title={Learning from Videos for 3D World: Enhancing MLLMs with 3D Vision Geometry Priors},
  author={Zheng, Duo and Huang, Shijia and Li, Yanyang and Wang, Liwei},
  journal={arXiv preprint arXiv:2505.24625},
  year={2025}
}

@article{xia2025drivelaw,
  title={DriveLaW: Unifying Planning and Video Generation in a Latent Driving World},
  author={Xia, Tianze and Li, Yongkang and Zhou, Lijun and Yao, Jingfeng and Xiong, Kaixin and Sun, Haiyang and Wang, Bing and Ma, Kun and Ye, Hangjun and Liu, Wenyu and others},
  journal={arXiv preprint arXiv:2512.23421},
  year={2025}
}

@article{li2025spatial,
  title={Spatial forcing: Implicit spatial representation alignment for vision-language-action model},
  author={Li, Fuhao and Song, Wenxuan and Zhao, Han and Wang, Jingbo and Ding, Pengxiang and Wang, Donglin and Zeng, Long and Li, Haoang},
  journal={arXiv preprint arXiv:2510.12276},
  year={2025}
}

@article{lou2025adacot,
  title={AdaCoT: Pareto-Optimal Adaptive Chain-of-Thought Triggering via Reinforcement Learning},
  author={Lou, Chenwei and Sun, Zewei and Liang, Xinnian and Qu, Meng and Shen, Wei and Wang, Wenqi and Li, Yuntao and Yang, Qingping and Wu, Shuangzhi},
  journal={arXiv preprint arXiv:2505.11896},
  year={2025}
}

@article{zhang2025adaptthink,
  title={Adaptthink: Reasoning models can learn when to think},
  author={Zhang, Jiajie and Lin, Nianyi and Hou, Lei and Feng, Ling and Li, Juanzi},
  journal={arXiv preprint arXiv:2505.13417},
  year={2025}
}

@article{jiang2025alphadrive,
  title={Alphadrive: Unleashing the power of vlms in autonomous driving via reinforcement learning and reasoning},
  author={Jiang, Bo and Chen, Shaoyu and Zhang, Qian and Liu, Wenyu and Wang, Xinggang},
  journal={arXiv preprint arXiv:2503.07608},
  year={2025}
}

@article{zhu2026analyzing,
  title={Analyzing Reasoning Consistency in Large Multimodal Models under Cross-Modal Conflicts},
  author={Zhu, Zhihao and Liang, Jiafeng and Jiang, Shixin and Fu, Jinlan and Liu, Ming and Sun, Guanglu and Ng, See-Kiong and Qin, Bing},
  journal={arXiv preprint arXiv:2601.04073},
  year={2026}
}

@inproceedings{wang2025vggt,
  title={Vggt: Visual geometry grounded transformer},
  author={Wang, Jianyuan and Chen, Minghao and Karaev, Nikita and Vedaldi, Andrea and Rupprecht, Christian and Novotny, David},
  booktitle={Proceedings of the Computer Vision and Pattern Recognition Conference},
  pages={5294--5306},
  year={2025}
}

@article{wang2024drivecot,
  title={Drivecot: Integrating chain-of-thought reasoning with end-to-end driving},
  author={Wang, Tianqi and Xie, Enze and Chu, Ruihang and Li, Zhenguo and Luo, Ping},
  journal={arXiv preprint arXiv:2403.16996},
  year={2024}
}

@article{hao2024training,
  title={Training large language models to reason in a continuous latent space},
  author={Hao, Shibo and Sukhbaatar, Sainbayar and Su, DiJia and Li, Xian and Hu, Zhiting and Weston, Jason and Tian, Yuandong},
  journal={arXiv preprint arXiv:2412.06769},
  year={2024}
}

@article{ray2025mull,
  title={Mull-Tokens: Modality-Agnostic Latent Thinking},
  author={Ray, Arijit and Abdelkader, Ahmed and Mao, Chengzhi and Plummer, Bryan A and Saenko, Kate and Krishna, Ranjay and Guibas, Leonidas and Chu, Wen-Sheng},
  journal={arXiv preprint arXiv:2512.10941},
  year={2025}
}

@article{wang2025alpamayo,
  title={Alpamayo-r1: Bridging reasoning and action prediction for generalizable autonomous driving in the long tail},
  author={Wang, Yan and Luo, Wenjie and Bai, Junjie and Cao, Yulong and Che, Tong and Chen, Ke and Chen, Yuxiao and Diamond, Jenna and Ding, Yifan and Ding, Wenhao and others},
  journal={arXiv preprint arXiv:2511.00088},
  year={2025}
}

@article{tian2024drivevlm,
  title={Drivevlm: The convergence of autonomous driving and large vision-language models},
  author={Tian, Xiaoyu and Gu, Junru and Li, Bailin and Liu, Yicheng and Wang, Yang and Zhao, Zhiyong and Zhan, Kun and Jia, Peng and Lang, Xianpeng and Zhao, Hang},
  journal={arXiv preprint arXiv:2402.12289},
  year={2024}
}

@article{hwang2024emma,
  title={Emma: End-to-end multimodal model for autonomous driving},
  author={Hwang, Jyh-Jing and Xu, Runsheng and Lin, Hubert and Hung, Wei-Chih and Ji, Jingwei and Choi, Kristy and Huang, Di and He, Tong and Covington, Paul and Sapp, Benjamin and others},
  journal={arXiv preprint arXiv:2410.23262},
  year={2024}
}

@inproceedings{wang2025omnidrive,
  title={Omnidrive: A holistic vision-language dataset for autonomous driving with counterfactual reasoning},
  author={Wang, Shihao and Yu, Zhiding and Jiang, Xiaohui and Lan, Shiyi and Shi, Min and Chang, Nadine and Kautz, Jan and Li, Ying and Alvarez, Jose M},
  booktitle={Proceedings of the Computer Vision and Pattern Recognition Conference},
  pages={22442--22452},
  year={2025}
}

@article{fu2025orion,
  title={Orion: A holistic end-to-end autonomous driving framework by vision-language instructed action generation},
  author={Fu, Haoyu and Zhang, Diankun and Zhao, Zongchuang and Cui, Jianfeng and Liang, Dingkang and Zhang, Chong and Zhang, Dingyuan and Xie, Hongwei and Wang, Bing and Bai, Xiang},
  journal={arXiv preprint arXiv:2503.19755},
  year={2025}
}

@inproceedings{hu2023planning,
  title={Planning-oriented autonomous driving},
  author={Hu, Yihan and Yang, Jiazhi and Chen, Li and Li, Keyu and Sima, Chonghao and Zhu, Xizhou and Chai, Siqi and Du, Senyao and Lin, Tianwei and Wang, Wenhai and others},
  booktitle={Proceedings of the IEEE/CVF conference on computer vision and pattern recognition},
  pages={17853--17862},
  year={2023}
}

@inproceedings{jiang2023vad,
  title={Vad: Vectorized scene representation for efficient autonomous driving},
  author={Jiang, Bo and Chen, Shaoyu and Xu, Qing and Liao, Bencheng and Chen, Jiajie and Zhou, Helong and Zhang, Qian and Liu, Wenyu and Huang, Chang and Wang, Xinggang},
  booktitle={Proceedings of the IEEE/CVF International Conference on Computer Vision},
  pages={8340--8350},
  year={2023}
}

@article{li2025recogdrive,
  title={ReCogDrive: A Reinforced Cognitive Framework for End-to-End Autonomous Driving},
  author={Li, Yongkang and Xiong, Kaixin and Guo, Xiangyu and Li, Fang and Yan, Sixu and Xu, Gangwei and Zhou, Lijun and Chen, Long and Sun, Haiyang and Wang, Bing and others},
  journal={arXiv preprint arXiv:2506.08052},
  year={2025}
}

@article{zhou2025autovla,
  title={AutoVLA: A Vision-Language-Action Model for End-to-End Autonomous Driving with Adaptive Reasoning and Reinforcement Fine-Tuning},
  author={Zhou, Zewei and Cai, Tianhui and Zhao, Seth Z and Zhang, Yun and Huang, Zhiyu and Zhou, Bolei and Ma, Jiaqi},
  journal={arXiv preprint arXiv:2506.13757},
  year={2025}
}

@article{shao2024deepseekmath,
  title={Deepseekmath: Pushing the limits of mathematical reasoning in open language models},
  author={Shao, Zhihong and Wang, Peiyi and Zhu, Qihao and Xu, Runxin and Song, Junxiao and Bi, Xiao and Zhang, Haowei and Zhang, Mingchuan and Li, YK and Wu, Yang and others},
  journal={arXiv preprint arXiv:2402.03300},
  year={2024}
}

@article{zhu2025internvl3,
  title={Internvl3: Exploring advanced training and test-time recipes for open-source multimodal models},
  author={Zhu, Jinguo and Wang, Weiyun and Chen, Zhe and Liu, Zhaoyang and Ye, Shenglong and Gu, Lixin and Tian, Hao and Duan, Yuchen and Su, Weijie and Shao, Jie and others},
  journal={arXiv preprint arXiv:2504.10479},
  year={2025}
}

@article{chitta2022transfuser,
  title={Transfuser: Imitation with transformer-based sensor fusion for autonomous driving},
  author={Chitta, Kashyap and Prakash, Aditya and Jaeger, Bernhard and Yu, Zehao and Renz, Katrin and Geiger, Andreas},
  journal={IEEE transactions on pattern analysis and machine intelligence},
  volume={45},
  number={11},
  pages={12878--12895},
  year={2022},
  publisher={IEEE}
}

@article{li2024hydra,
  title={Hydra-mdp: End-to-end multimodal planning with multi-target hydra-distillation},
  author={Li, Zhenxin and Li, Kailin and Wang, Shihao and Lan, Shiyi and Yu, Zhiding and Ji, Yishen and Li, Zhiqi and Zhu, Ziyue and Kautz, Jan and Wu, Zuxuan and others},
  journal={arXiv preprint arXiv:2406.06978},
  year={2024}
}

@inproceedings{liao2025diffusiondrive,
  title={Diffusiondrive: Truncated diffusion model for end-to-end autonomous driving},
  author={Liao, Bencheng and Chen, Shaoyu and Yin, Haoran and Jiang, Bo and Wang, Cheng and Yan, Sixu and Zhang, Xinbang and Li, Xiangyu and Zhang, Ying and Zhang, Qian and others},
  booktitle={Proceedings of the Computer Vision and Pattern Recognition Conference},
  pages={12037--12047},
  year={2025}
}

@article{li2025end,
  title={End-to-end driving with online trajectory evaluation via bev world model},
  author={Li, Yingyan and Wang, Yuqi and Liu, Yang and He, Jiawei and Fan, Lue and Zhang, Zhaoxiang},
  journal={arXiv preprint arXiv:2504.01941},
  year={2025}
}

@article{dauner2024navsim,
  title={Navsim: Data-driven non-reactive autonomous vehicle simulation and benchmarking},
  author={Dauner, Daniel and Hallgarten, Marcel and Li, Tianyu and Weng, Xinshuo and Huang, Zhiyu and Yang, Zetong and Li, Hongyang and Gilitschenski, Igor and Ivanovic, Boris and Pavone, Marco and others},
  journal={Advances in Neural Information Processing Systems},
  volume={37},
  pages={28706--28719},
  year={2024}
}

@inproceedings{sima2024drivelm,
  title={Drivelm: Driving with graph visual question answering},
  author={Sima, Chonghao and Renz, Katrin and Chitta, Kashyap and Chen, Li and Zhang, Hanxue and Xie, Chengen and Bei{\ss}wenger, Jens and Luo, Ping and Geiger, Andreas and Li, Hongyang},
  booktitle={European conference on computer vision},
  pages={256--274},
  year={2024},
  organization={Springer}
}

@article{bai2025qwen2,
  title={Qwen2. 5-vl technical report},
  author={Bai, Shuai and Chen, Keqin and Liu, Xuejing and Wang, Jialin and Ge, Wenbin and Song, Sibo and Dang, Kai and Wang, Peng and Wang, Shijie and Tang, Jun and others},
  journal={arXiv preprint arXiv:2502.13923},
  year={2025}
}

@article{li2025drivevla,
  title={DriveVLA-W0: World Models Amplify Data Scaling Law in Autonomous Driving},
  author={Li, Yingyan and Shang, Shuyao and Liu, Weisong and Zhan, Bing and Wang, Haochen and Wang, Yuqi and Chen, Yuntao and Wang, Xiaoman and An, Yasong and Tang, Chufeng and others},
  journal={arXiv preprint arXiv:2510.12796},
  year={2025}
}

@article{cao2025pseudo,
  title={Pseudo-simulation for autonomous driving},
  author={Cao, Wei and Hallgarten, Marcel and Li, Tianyu and Dauner, Daniel and Gu, Xunjiang and Wang, Caojun and Miron, Yakov and Aiello, Marco and Li, Hongyang and Gilitschenski, Igor and others},
  journal={arXiv preprint arXiv:2506.04218},
  year={2025}
}

@article{drivemllm,
  title={SURDS: Benchmarking Spatial Understanding and Reasoning in Driving Scenarios with Vision Language Models},
  author={Guo, Xianda and Zhang, Ruijun and Duan, Yiqun and He, Yuhang and Nie, Dujun and Huang, Wenke and Zhang, Chenming and Liu, Shuai and Zhao, Hao and Chen, Long},
  journal={arXiv preprint arXiv:2411.13112},
  year={2024}
}

@article{agarwal2025cosmos,
  title={Cosmos world foundation model platform for physical ai},
  author={Agarwal, Niket and Ali, Arslan and Bala, Maciej and Balaji, Yogesh and Barker, Erik and Cai, Tiffany and Chattopadhyay, Prithvijit and Chen, Yongxin and Cui, Yin and Ding, Yifan and others},
  journal={arXiv preprint arXiv:2501.03575},
  year={2025}
}

@article{zeng2025futuresightdrive,
  title={Futuresightdrive: Thinking visually with spatio-temporal cot for autonomous driving},
  author={Zeng, Shuang and Chang, Xinyuan and Xie, Mengwei and Liu, Xinran and Bai, Yifan and Pan, Zheng and Xu, Mu and Wei, Xing and Guo, Ning},
  journal={arXiv preprint arXiv:2505.17685},
  year={2025}
}

@article{zhou2025opendrivevla,
  title={Opendrivevla: Towards end-to-end autonomous driving with large vision language action model},
  author={Zhou, Xingcheng and Han, Xuyuan and Yang, Feng and Ma, Yunpu and Tresp, Volker and Knoll, Alois},
  journal={arXiv preprint arXiv:2503.23463},
  year={2025}
}

@article{zheng2024doe,
  title={Doe-1: Closed-loop autonomous driving with large world model},
  author={Zheng, Wenzhao and Xia, Zetian and Huang, Yuanhui and Zuo, Sicheng and Zhou, Jie and Lu, Jiwen},
  journal={arXiv preprint arXiv:2412.09627},
  year={2024}
}

@inproceedings{huang2024making,
  title={Making large language models better planners with reasoning-decision alignment},
  author={Huang, Zhijian and Tang, Tao and Chen, Shaoxiang and Lin, Sihao and Jie, Zequn and Ma, Lin and Wang, Guangrun and Liang, Xiaodan},
  booktitle={European Conference on Computer Vision},
  pages={73--90},
  year={2024},
  organization={Springer}
}

@inproceedings{caesar2020nuscenes,
  title={nuscenes: A multimodal dataset for autonomous driving},
  author={Caesar, Holger and Bankiti, Varun and Lang, Alex H and Vora, Sourabh and Liong, Venice Erin and Xu, Qiang and Krishnan, Anush and Pan, Yu and Baldan, Giancarlo and Beijbom, Oscar},
  booktitle={Proceedings of the IEEE/CVF conference on computer vision and pattern recognition},
  pages={11621--11631},
  year={2020}
}
